\documentclass{article}

\usepackage{microtype}
\usepackage{graphicx}
\usepackage{subfigure}
\usepackage{booktabs} 

\usepackage{hyperref}

\usepackage[table]{xcolor}

\usepackage{color,soul}

\usepackage[accepted]{icml2024}

\usepackage{amsmath}
\usepackage{amssymb}
\usepackage{mathtools}
\usepackage{amsthm}

\usepackage[capitalize,noabbrev]{cleveref}

\usepackage{graphics}

\theoremstyle{plain}

\theoremstyle{definition}

\theoremstyle{remark}

\usepackage[textsize=tiny]{todonotes}

\begin{document}

\twocolumn[
\icmltitle{Hierarchical Neural Operator Transformer with Learnable Frequency-aware Loss Prior for Arbitrary-scale Super-resolution}



\icmlsetsymbol{equal}{*}

\begin{icmlauthorlist}
\icmlauthor{Xihaier Luo}{bnl}
\icmlauthor{Xiaoning Qian}{bnl,tamu}
\icmlauthor{Byung-Jun Yoon}{bnl,tamu}
\end{icmlauthorlist}

\icmlaffiliation{bnl}{Computational Science Intiative, Brookhaven National Laboratory, Upton, NY, USA}
\icmlaffiliation{tamu}{Department of Electrical and Computer Engineering, Texas A\&M University, College Station, TX, USA}

\icmlcorrespondingauthor{Xihaier Luo}{xluo@bnl.gov}
\icmlcorrespondingauthor{Byung-Jun Yoon}{bjyoon@tamu.edu}

\icmlkeywords{arbitrary-scale super-resolution, hierarchical neural operator, frequency-aware loss}

\vskip 0.3in
]



\printAffiliationsAndNotice{}  

\begin{abstract}
In this work, we present an \textit{arbitrary-scale} super-resolution (SR) method to enhance the resolution of scientific data, which often involves complex challenges such as continuity, multi-scale physics, and the intricacies of high-frequency signals. Grounded in operator learning, the proposed method is resolution-invariant. The core of our model is a hierarchical neural operator that leverages a Galerkin-type self-attention mechanism, enabling efficient learning of mappings between function spaces. Sinc filters are used to facilitate the information transfer across different levels in the hierarchy, thereby ensuring representation equivalence in the proposed neural operator. Additionally, we introduce a learnable prior structure that is derived from the spectral resizing of the input data. This loss prior is model-agnostic and is designed to dynamically adjust the weighting of pixel contributions, thereby balancing gradients effectively across the model. We conduct extensive experiments on diverse datasets from different domains and demonstrate consistent improvements compared to strong baselines, which consist of various state-of-the-art SR methods.
\end{abstract}

\section{Introduction}
\label{intro}

Super-resolution (SR) plays a pivotal role in low-level vision tasks. The primary objective of SR is to transform blurred, fuzzy, and low-resolution images into clear, high-resolution images with enhanced visual perception. In recent years, deep learning has significantly advanced SR and has demonstrated promising performances in diverse domains beyond computer vision, including but not limited to medical imaging~\citep{li2022transformer}, climate modeling~\citep{reichstein2019deep}, and remote sensing~\citep{akiva2022self}. Nevertheless, existing deep learning-based SR methods often limit themselves to a \textit{fixed-scale} (e.g., $\times2, \times3, \times4$). The emergence of implicit neural representation (INR) in computer vision allows for continuous representation of complex 2D/3D objects and scenes~\citep{xie2022neural}. This development introduces opportunities for \textit{arbitrary-scale} SR.

\textbf{Challenges}: Current \textit{arbitrary-scale} SR methods, while capable of learning continuous representations from discretized data, face several challenges. 
1) \textit{Low-resolution Features}: The spatial resolution of extracted features is often inadequate for dense tasks such as image segmentation and regression. For example, using ResNet-50 on a $224 \times 224$ pixel input results in $7 \times 7$ deep features, showing a marked loss of resolution due to aggressive pooling~\citep{he2016deep}. Even without resolution reduction, empirical evidence shows performance decline after flipping the feature map-a phenomenon termed flipping consistency decline~\citep{song2023ope}.
2) \textit{Spectral Bias}: INRs are coordinate-based continuous functions usually parameterized by a multi-layer perceptron (MLP). The point-wise behavior of MLP in spatial dimensions poses challenges in learning high-frequency information, commonly known as spectral bias~\citep{rahaman2019spectral}. This issue is particularly problematic when modeling scientific data, where the super-resolved predictions often appear over-diffused and fail to capture fine-scale details, such as small-scale vortices in turbulent flows~\citep{fukami2019super}. 
3) \textit{Loss Function}: 
Most arbitrary-scale methods rely on per-pixel loss metrics (e.g., L1/L2 loss)~\citep{liu2023arbitrary}. Training the model with a per-pixel L1 loss in a regression manner biases the reconstruction error towards an averaged output of all potential high-resolution images. Consequently, this often leads to blurry model predictions.

\textbf{Solutions}:
For challenge 1), the key solution lies in upsampling deep features. Traditional spatial upsampling methods, such as transposed convolution, focus on local pixel attention and overlook global dependencies~\citep{dumoulin2016guide}. In contrast, Fourier domain upsampling supports global modeling. Therefore, we propose a hybrid approach that combines traditional convolution with spectral upsampling for enhanced performance. For challenge 2), we reformulate SR as operator learning~\citep{kovachki2023neural}. We replace the MLP-based inference network in an INR with a neural operator. This approach considers images as continuous functions rather than 2D pixel arrays, with each image instance representing a discretization of an underlying function. Here, we propose a hierarchical transformer as a neural operator for learning mappings between these function spaces.
For challenge 3), we propose a spectral resizing-based loss prior. This prior is designed to adjust the contribution of each pixel to the overall loss within the image space. Such re-weighting redistributes gradients, thereby improving the SR model's ability to capture details across both high- and low-frequency regions.

The major contributions of our work include:
\begin{itemize}
    \vspace{-2.7mm}
    \item We introduce a new hierarchical neural operator based on the transformer architecture with a Galerkin-type self-attention mechanism for \textit{arbitrary-scale} super-resolution of scientific data.
    \item We devise a simple yet highly effective mechanism to construct a loss prior. It strategically adjusts pixel loss contributions during training and rebalances them in the next gradient updating step.
    \item We carry out extensive experiments to assess the effectiveness of our proposed method, showcasing its superiority compared to existing state-of-the-art (SOTA) SR approaches.
\end{itemize}

\section{Background and Related Work}
\label{Background}
Super-resolution in many cases is a challenging ill-posed inverse problem. Its primary goal is to establish a mapping $\mathcal{F}: \mathbb{R}^{\mathbb{D}_{a}} \rightarrow \mathbb{R}^{\mathbb{D}_{b}}$ that transforms a given low-resolution~(LR) input, denoted as $\boldsymbol{a}$, into a high-resolution~(HR) output, denoted as $\boldsymbol{b}$, where $\mathbb{D}$ represents the discretization function and $\mathbb{D}(\boldsymbol{a})$ represents the discretized resolution of $\boldsymbol{a}$. The SR problem has traditionally been addressed by a spectrum of techniques, such as interpolation~\citep{keys1981cubic}, neighbor embedding~\citep{chang2004super}, sparse coding~\citep{yang2010image}, and dictionary-based learning~\citep{timofte2013anchored}. Overall, these approaches struggle to capture the intricate and nonlinear LR-to-HR transformation, often resulting in subpar super-resolved images, particularly in contexts involving high-wavenumbers (e.g. turbulence in~\citet{xie2018tempogan} and climate in~\citet{stengel2020adversarial}). On the other hand, the success of deep learning has brought about significant advancements in this field.

\subsection{Deep Super-Resolution Models}
\label{dsrm}

\textbf{Single-scale SR.} \citet{dong2014learning} were the first to introduce the CNN-based network to address SR challenges in natural images, achieving superior results compared to conventional methods. Following this, VDSR incorporated residual blocks~\citep{kim2016accurate}, EDSR removed batch normalization layers, utilizing a residual scaling technique for training~\citep{lim2017enhanced}, RDN introduced the concept of dense feature fusion~\citep{zhang2018residual}, and SwinIR proposed bi-level feature extraction mechanism for local attention and cross-window interaction~\citep{liang2021swinir}. These advancements collectively contributed to further enhancing the performance of super-resolution methods. Nonetheless, these approaches are confined to conducting upsampling with predefined factors, necessitating the training of separate models for each upsampling ratio, e.g. $\mathcal{F}_1 $ for $ \mathbb{D}(\boldsymbol{a}) \xrightarrow{s_1} \mathbb{D}(\boldsymbol{b})$ and $\mathcal{F}_2 $ for $ \mathbb{D}(\boldsymbol{a}) \xrightarrow{s_2} \mathbb{D}(\boldsymbol{b})$. This constraint restricts the practical applicability of these \textit{single-scale} SR models~\citep{liu2023arbitrary}.

\textbf{Arbitrary-scale SR.} Addressing this issue, a more recent and practical approach of \textit{arbitrary-scale} super-resolution has emerged. MetaSR provides the first single-model solution for \textit{arbitrary-scale} SR by predicting convolutional filter weights based on scale factors and coordinates, enabling adaptive filter weight prediction~\citep{hu2019meta}. In contrast to Meta-SR, LIIF employs an MLP as a local implicit function to predict RGB values based on queried HR image coordinates, extracted LR image features, and a cell size~\citep{chen2021learning}. LTE further introduced a local texture estimator that converts coordinates into Fourier domain data, enhancing the representational capacity of its local implicit function~\citep{lee2022local}. More recently, LINF~\citep{yao2023local} and CiaoSR~\citep{cao2023ciaosr} have emerged to enhance LIIF's implicit neural representation. LINF focuses on learning local texture patch distributions, employing coordinate conditional normalizing flow for conditional signal generation. Alternatively, CiaoSR introduces an implicit attention network to determine ensemble weights for nearby local features. While effective, many \textit{arbitrary-scale} SR methods heavily depend on continuous functions parameterized by MLPs. Unfortunately, MLPs are recognized for struggling with the learning of high-frequency functions, limiting their applicability in scientific domains characterized by complex, multiphysics-multiscale processes, such as weather data~\citep{luo2023reinstating}.

\begin{figure*}
\begin{center}
\centerline{\includegraphics[width=2\columnwidth]{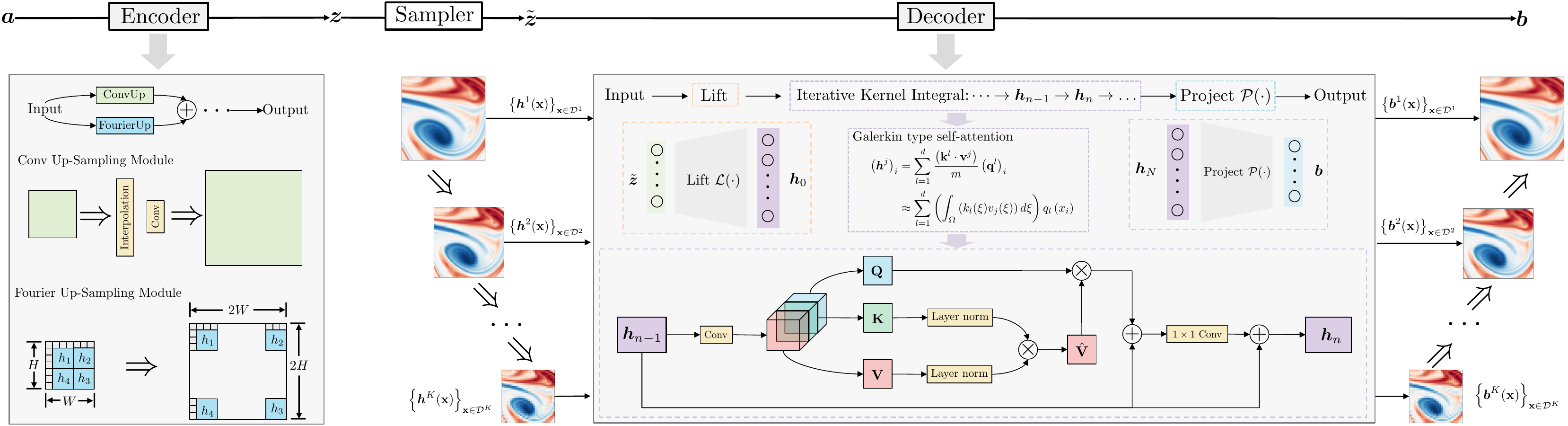}}
\caption{Overview of the \textbf{Hi}erarchical \textbf{N}eural \textbf{O}perator \textbf{T}ransform\textbf{E}r (HiNOTE). HiNOTE features a structured architecture comprising three key components: Firstly, an encoder designed for upsampling deep features; secondly, a sampler tasked with rendering a specific set of features; and thirdly, a decoder capable of making inferences at various arbitrary points within the domain.}
\label{fig:model}
\end{center}
\vskip -0.3in
\end{figure*}

\subsection{Neural Operator}
\label{no}
Recently, neural operators have arisen as a promising approach for approximating functions~\citep{kovachki2023neural}. Unlike fully connected neural networks and convolutional neural networks, neural operators incorporate function space characteristics to guide network training, demonstrating superior performance in nonlinear fitting and maintaining invariance through discretization~\citep{lu2021learning,li2021fourier,hao2023gnot,tran2023factorized}. SRNO exemplifies the application of neural operators in super-resolution tasks, employing a linear attention operator for Petrov-Galerkin projection~\citep{wei2023super}. DFNO is another instance employing a Fourier neural operator for downscaling climate variables~\citep{yang2023fourier}. This model undergoes training with data featuring a modest upsampling factor and subsequently demonstrates the ability to perform zero-shot downscaling to arbitrary, unseen high resolutions.


\section{Method}
\label{method}
To achieve \textit{arbitrary-scale} SR, we introduce \textbf{Hi}erarchical \textbf{N}eural \textbf{O}perator \textbf{T}ransform\textbf{E}r (HiNOTE), featuring a hybrid upsampling module and a frequency-aware loss prior (See Fig.~\ref{fig:model}). These designs enable accurate learning of the continuous representation of the underlying data.

\subsection{Problem Statement}

\textbf{Problem Reformulation.} In the context of single-scale SR, different methods aim to establish a mapping $f: \mathbb{R}^{d_a} \rightarrow \mathbb{R}^{d_b}$ from low-resolution input $\boldsymbol{a} \in \mathbb{R}^{d_a}$ to high-resolution output $\boldsymbol{b} \in \mathbb{R}^{d_b}$ with $d_a < d_b$. This formulation imposes a constraint fixing the input/output resolution at $d_a/d_b$. To overcome this limitation, we propose to learn a mapping from an infinite-dimensional function space to another infinite-dimensional function space $\mathcal{G}^\dagger: \mathcal{A} \rightarrow \mathcal{B}$. Here, $\mathcal{A}=\mathcal{A}(D;\mathbb{R}^{d_a})$ and $\mathcal{B}=\mathcal{B}(D;\mathbb{R}^{d_b})$ are separable Banach spaces of functions taking values in $\mathbb{R}^{d_a}$ and $\mathbb{R}^{d_b}$ respectively~\citep{kovachki2023neural}. This approach models observed digital data (dynamic fields, images of 2D pixel arrays, or 1D acoustic signals) by continuous function representations, treating corresponding data as observed measurements in discretized domains of the underlying functions. It allows for obtaining arbitrarily low-resolution inputs and high-resolution outputs by evaluating $\boldsymbol{a}$ and $\boldsymbol{b}$ at numerous points within $D$. For instance, $\boldsymbol{a} (\mathbb{D}) = \{a(x_1), ..., a(x_n)\}$, where $\mathbb{D} = \{ x_1, \dots, x_n \} \subset D$ is a $n$-point discretization of the domain D. For simplicity in the following discussion, $\boldsymbol{a}$ and $\boldsymbol{b}$ will be used unless stated otherwise.

\textbf{Optimization Goal.} This new formulation based on continuous function representations enables arbitrary-scale SR: Given observations $\{\boldsymbol{a}^{(j)}, \boldsymbol{b}^{(j)}\}_{j=1}^N$, where $\boldsymbol{a}^{(j)}$ is an i.i.d. low-resolution sample and $\boldsymbol{b}^{(j)} = \mathcal{G}^\dagger(\boldsymbol{a}^{(j)})$ represents the high-resolution counterpart, we seek to build a parametric map to approximate $\mathcal{G}^\dagger$: 
\begin{equation}
\label{eq:prob}
\mathcal{G}_{\theta}:\mathcal{A} \times \Theta \rightarrow \mathcal{B}, \quad \theta \in \Theta
\end{equation} 
for some finite-dimensional parameter space $\Theta$ by choosing $\theta^{\dagger} \in \Theta$ so that $\mathcal{G}_{\theta^{\dagger}} \approx \mathcal{G}^\dagger$. 

\subsection{Overview of Model Architecture}

Following the defined problem, to achieve the goal of learning the mapping from the function space $\mathcal{A}$ to $\mathcal{B}$, we design $\mathcal{G}_{\theta}$ as a composition of a hybrid upsampling-based encoder, $E_\varphi$, and more importantly, a new hierarchical neural operator transformer-based decoder, $D_\phi$, stacked before and after a parameter-free sampler $\mathcal{S}$ (Fig.~\ref{fig:model}): 
\begin{equation}
\mathcal{G}_\theta: = D_\phi \circ \mathcal{S} \circ E_\varphi
\end{equation}

\begin{enumerate}
    \item The encoder $E_\varphi: \mathcal{A}(D; \mathbb{R}^{d_a}) \rightarrow \mathcal{Z}(D; \mathbb{R}^{d_z})$ is a functional operator parameterized by $\varphi$. It maps the input data $\boldsymbol{a}^{(j)}$ to a condensed representation, i.e., a feature vector $\boldsymbol{z}^{(j)} = E(\boldsymbol{a}^{(j)}; \varphi)$. We incorporate a hybrid upsampling module into model $E_\varphi$, designed to effectively capture both spatial and spectral information as well as local and global details.
    \item The sampler $\mathcal{S}$ is parameter-free and acts as an up-sampling module $\mathcal{S}: \mathcal{Z}(D; \mathbb{R}^{d_z}) \rightarrow \mathcal{Z}(D; \mathbb{R}^{d_b})$. It aligns the encoder output size $d_z$ with any arbitrarily chosen high-resolution discretization $d_b$.
    \item The decoder $D_\phi: \mathcal{Z}(D; \mathbb{R}^{d_b}) \rightarrow \mathcal{B}(D; \mathbb{R}^{d_b})$ is a functional operator parameterized by $\phi$. It enables the generation of super-resolved outputs at any specified resolution during the inference $\boldsymbol{b}^{(j)} = D(\boldsymbol{z}^{(j)}; \phi)$.
\end{enumerate}

\subsection{Encoder}
\label{encoder}
The encoder is engineered to extract a series of deep features from the input data. However, these deep features often lack the spatial resolution needed for tasks like segmentation and depth prediction. This spatial insufficiency is amplified in the SR setting, where every pixel from low-resolution input is important. Common spatial up-sampling methods in convolutional neural networks rely on local pixel attention, limiting global dependency exploration. In contrast, the Fourier domain aligns with global modeling principles, as per the spectral convolution theorem~\citep{brigham1988fast}. To address this, we introduce a hybrid upsampling module (See the encoder details in Fig.~\ref{fig:model}). It merges conventional convolution-based upsampling with a recently developed Fourier upsampling module~\citep{yu2022deep}. Our proposed hybrid upsampling module effectively captures global features and preserves overall structural integrity while also benefiting from the local context awareness inherent in convolutional operations. An additional enhancement we introduced pertains to the placement of the upsampling module. While conventional SR models typically position the upsampling layer towards the end of the network. We propose that refining feature maps at the network's onset is more advantageous (See Fig.~\ref{fig:improvment}). 


\subsection{Sampler}
\label{sampler}
\begin{figure}[ht]
\begin{center}
\centerline{\includegraphics[width=\columnwidth]{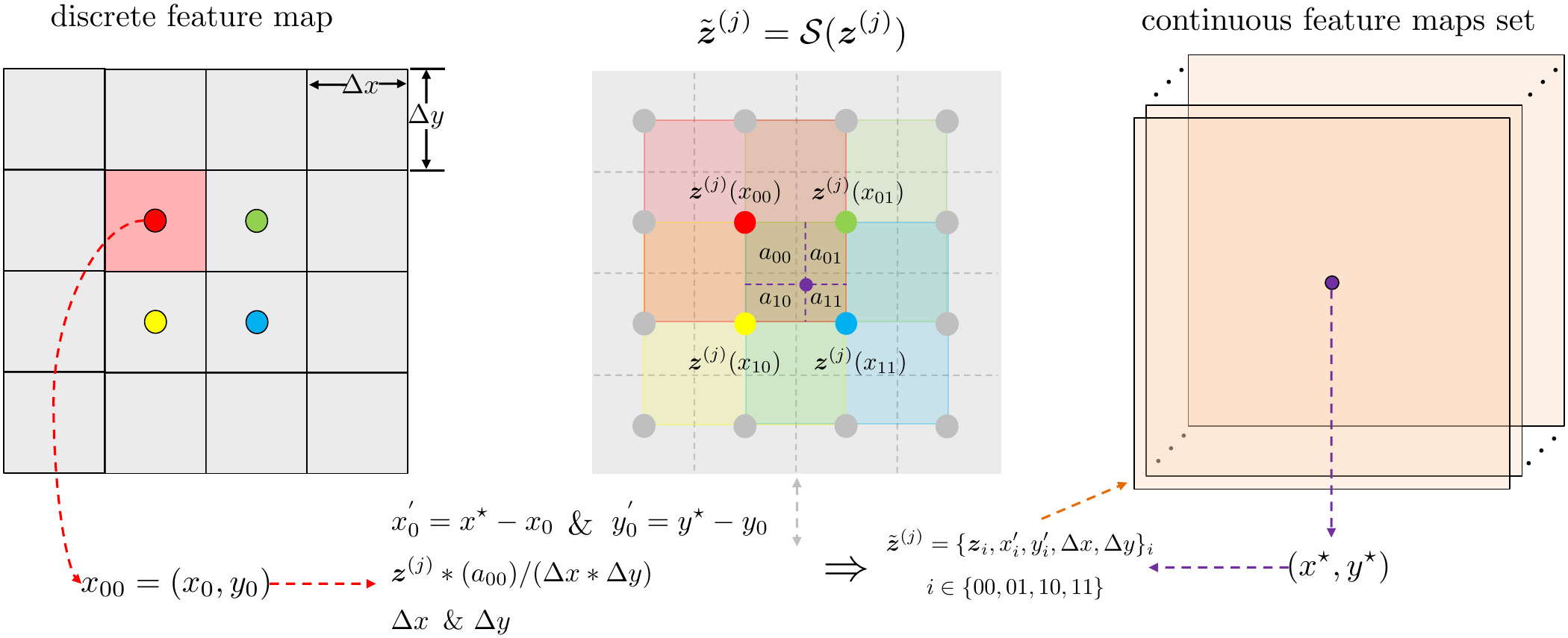}}
\caption{Illustration of the parameter-free sampler. It samples arbitrary resolutions from feature maps extracted by the encoder and combines the positional information of the grid points.}
\label{fig:upsampling}
\end{center}
\vskip -0.3in
\end{figure}

The sampler operator utilizes a patch-based interpolation scheme to transform a discrete feature vector from the existing resolution $d_z$ to any target resolution $d_b$. Specifically, this involves two steps: \textit{feature map rendering} and \textit{patch ensemble}, as shown in Fig.~\ref{fig:upsampling}.

\textbf{Feature Map Rendering.} For rendering a new feature map from a discrete feature map with size $h \times w$, we assume feature vectors are evenly distributed in a 2D domain $[-1, 1] \times [-1, 1]$. Dividing this into $h \times w$ regions, each cell ($\Delta x \times \Delta y$) in the feature map $\boldsymbol{z}^{(j)}$ is associated with absolute central coordinates $(x, y)$ in the corresponding region. Given an arbitrary point at $(x^{\star}, y^{\star})$, we first calculate the distances to the nearest four neighboring central coordinates $(x_i, y_i)$, where $i \in \{00, 01, 10, 11\}$ represents the nearest feature vectors in the top-left, top-right, bottom-left, and bottom-right sub-spaces: 
\begin{equation}
x_i^{\prime}=\frac{\left(x^{\star}-x_i\right) \cdot \Delta x}{2}, \quad y_i^{\prime}=\frac{\left(y^{\star}-y_i\right) \cdot \Delta y}{2}. 
\end{equation}

We then normalize the feature vectors $\boldsymbol{z}^{(j)} (x_i, y_i)$ based on the area of the rectangle $a_i$ between the query point and its nearest feature vector’s diagonal counterpart: 
\begin{equation}
\boldsymbol{z}^{(j)} (x^{\star}, y^{\star}) = \boldsymbol{z}^{(j)} (x_i, y_i) \cdot a_i (x_i^{\prime}, y_i^{\prime}) \, / \, (\Delta x \cdot \Delta y). 
\end{equation}

\textbf{Patch Ensemble.} Instead of merging normalized feature vectors, we propose feature ensembling. These normalized features, along with positional information, are fused in the decoder $D_\phi(\boldsymbol{z}^{(j)})$ to enhance local feature ensembling: 
\begin{equation}
\tilde{\boldsymbol{z}}^{(j)} = \{ \boldsymbol{z}^{(j)} (x^{\star}, y^{\star}), x_i, y_i, \Delta x, \Delta y \}_i. 
\end{equation}

Compared to direct weighted summation $\sum_{i} a_i \boldsymbol{z}^{(j)}_i$, which relies only on local feature coordinates~\citep{chen2021learning}, fusing continuous feature maps $\tilde{\boldsymbol{z}}^{(j)}$ in the decoder allows the model to leverage more information from local features.

\subsection{Decoder}
\label{no}
The decoder $D_\phi$ is a functional operator parameterized by a neural operator. It typically comprises three components: 
\begin{itemize}
  \item Lifting: The input $\tilde{\boldsymbol{z}} \in \mathcal{Z}$ undergoes lifting to its first hidden representation $\boldsymbol{h}_0 = \mathcal{L}(\Tilde{\boldsymbol{z}})$ using a pointwise function $\mathbb{R}^{d_{\Tilde{z}}} \rightarrow \mathbb{R}^{d_{h_0}}$. This lifting operation is achieved by a fully connected neural network with dimensions $d_{\Tilde{z}} < d_{h_0}$.
  \item Iterative Kernel Integral: For $n = 0, \dots , N-1$, each hidden representation evolves through an iterative kernel integral approximation $\left(\mathcal{K}_n\left(\boldsymbol{h}_n\right)\right)(x)=\int_{D} \kappa^{(n)}(x, y) \boldsymbol{h}_n(y) \mathrm{d}y, \, \forall x \in D$, where the kernel matrix $\kappa^{(n)}: \mathbb{R}^{d + d} \rightarrow \mathbb{R}^{d_h \times d_h}$ is a neural network.
  \item Projection: The output $\boldsymbol{u}$ is the projection of the last hidden representation's output, $\boldsymbol{b} = \mathcal{P}(\boldsymbol{h}_N)$, using a local transformation $\mathbb{R}^{d_{h_N}} \rightarrow \mathbb{R}^{d_b}$. Similar to the lifting step, this is performed by a fully connected neural network, typically with $d_{h_N} > d_b$.
\end{itemize}

The core of the decoder $D_\phi$ is the kernel integral, enabled by a transformer-based neural operator. This choice is driven by two primary factors: 
\begin{enumerate}
    \item \textit{Flexibility in Input and Output Sizes}: Transformers are adept at handling variable input and output data sizes~\citep{vaswani2017attention}. This adaptability is essential for our \textit{arbitrary-scale} SR tasks, where the domain size or discretization level may vary.
    \item \textit{Ability to Capture Long-Range Dependencies}: Transformers excel at capturing long-range dependencies in data, a critical feature for scientific applications like global weather pattern prediction, where interactions between distant variables are essential for accurate predictions~\citep{gao2022earthformer}.
\end{enumerate}

We complete the design of HiNote, our proposed framework. HiNOTE adopts a U-Net architecture with bandlimited functions as both inputs and outputs. This design renders HiNOTE a representationally equivalent neural operator in terms of aliasing errors~\citep{karras2021alias,raonic2023convolutional}. For the kernel integral, HiNOTE incorporates a Galerkin-type self-attention mechanism, which effectively reduces computational complexity from quadratic to linear~\citep{cao2021choose}.

\subsubsection{Hierarchical Architecture} 
Depicted in Fig.~\ref{fig:upsampling}, we use a parameterized downsample layer to derive deep representations $\{\{\boldsymbol{h}^k(\mathbf{x})\}_{\mathbf{x} \in \mathcal{D}^k}\}_{k=1}^K$ across $K$ scales. These representations aggregate local observations with learnable parameters, with $\left\{\boldsymbol{h}^1(\mathbf{x})\right\}_{\mathbf{x} \in \mathcal{D}^1}$ representing the finest resolution. Traditional downsampling methods, such as affine linear transformations combined with a nonlinear activation, often lead to aliasing errors by not adhering to the band-limits of the underlying function space~\citep{raonic2023convolutional}. To tackle this issue, we first upsample the input function beyond its frequency bandwidth; after the activation function, the signal is downsampled.

\textbf{Upsampling} The process involves initially augmenting the number of samples in the signal. For instance, consider upsampling a single channel signal $\boldsymbol{h}$ in $\mathbb{R}^{n_x \times n_y}$ to $\boldsymbol{h}^{\prime}$ in $\mathbb{R}^{n_x N \times n_y N}$. This is achieved by interspersing each pair of signal samples with $N-1$ zero-valued samples: 
\begin{equation}
\boldsymbol{h}^{\uparrow}[i, j]=\mathbb{I}_{n_x}(i) \cdot \mathbb{I}_{n_y}(j) \cdot \boldsymbol{h}[i \bmod n_x, j \bmod n_y], 
\end{equation} 
where $i=1 \ldots N \cdot n_x$ and $j=1 \ldots N \cdot n_y$. Subsequently, the upsampled signal is convolved with an interpolation filter, which serves to remove high-frequency components. 

\textbf{Downsampling} We utilize a sinc-based low-pass filter and execute downsampling post-nonlinear activation: 
\begin{equation}
\boldsymbol{h}^{\downarrow}  =\left(\frac{ w_{out}}{ w_{in}}\right)^2\left(f_{w_{out}} \star \boldsymbol{h}\right)(x), \quad \forall x \in D
\end{equation} 
where $\star$ is the convolution operation and $f_{w_{out}} (x_0, x_1) =\operatorname{sinc}\left(2 w x_0\right) \cdot \operatorname{sinc}\left(2 w x_1\right)$ is a sinc-based low-pass filter.

\subsubsection{Galerkin-type Self-attention}
Building upon the proposed hierarchical network structure, the next step is to conduct iterative kernel integral at each hierarchical level. Based on Nyström approximation theory, there is a similarity between the attention matrix in transformers and an integral kernel~\citep{kovachki2023neural}. More precisely, dot-product attention can be interpreted as an approximation of an integral transform using a non-symmetric, trainable kernel function~\citep{cao2021choose,li2023scalable}. In this study, we adopt the perspective of learnable kernel integrals for attention, treating each channel in the hidden feature map as a sample from a distinct function on the discretization grid. Omitting the layer index, let $\boldsymbol{h} = (h(x_1), \dots, h(x_m))^T \in \mathbb{R}^{m \times d_h}$ denote evaluations in the iterative kernel integral. Consider matrices $\{\mathbf{Q}, \mathbf{K}, \mathbf{V}\} \in \mathbb{R}^{m \times d_h}$ as query/key/value matrices. The columns of $\mathbf{Q} / \mathbf{K} / \mathbf{V}$ contain vector representations of learned basis functions, spanning subspaces in latent representation Hilbert spaces $\mathbf{Z}=\frac{1}{m} \mathbf{Q}\left(\widehat{\mathbf{K}}^T \widehat{\mathbf{V}}\right)$, where $\widehat{\cdot}$ denotes a column-wise normalized matrix. For instance, $\widehat{\mathbf{V}}_{ij}$ (also the \(i\)-th element of the \(j\)-th column vector: \( (\mathbf{v}^j)_i \)) is the evaluation of the \(j\)-th basis function on the \(i\)-th grid point \(x_i\), i.e., \( \widehat{\mathbf{V}}_{ij} = v_j(x_i)\). Similarly, for matrices $\mathbf{Q}, \mathbf{K}$, each column represents the sampling of basis functions $q_j(\cdot)$ and $k_j(\cdot)$, respectively. Leveraging this interpretation of learnable bases, we can employ the Monte-Carlo method, 
\begin{equation}
\left(\boldsymbol{h}^j\right)_i =\sum_{l=1}^d \frac{\left(\mathbf{k}^l \cdot \mathbf{v}^j\right)}{m}\left(\mathbf{q}^l\right)_i, 
\end{equation} 
to approximate the kernel integral. Hence, the kernel integral is iteratively executed through Galerkin-type self-attention~\citep{cao2021choose}. In contrast to standard attention, Galerkin-type self-attention reduces the quadratic complexity from $\mathcal{O}(m^2 d_h)$ to a linear complexity of $\mathcal{O}(m d^2_h)$.

\subsection{Training} 
\label{train}
Though our model $\mathcal{G}_\theta$ integrates three components $E_\varphi$, $\mathcal{S}$ and $D_\phi$, it is trained jointly in an end-to-end fashion. While DNN-based model excels in capturing complex signals, accurately approximating high-frequency details remains challenging~\citep{sitzmann2020implicit}. 

To address this, we introduce a frequency-aware loss prior in the corresponding discretized domain: 
\begin{equation}
\boldsymbol{p} = \left|\mathcal{G} (\boldsymbol{a}) - \mathcal{R}(\boldsymbol{a})\right|, \quad \boldsymbol{p} \in \mathbb{R}^{d_b}, 
\end{equation} 
where $\mathcal{R}: \mathbb{R}^{d_a} \rightarrow \mathbb{R}^{d_b}$ is the spectral resizing function. Next, we rescale $\boldsymbol{p}$ using min-max normalization $n(\boldsymbol{p}) = (\boldsymbol{p} - \text{min}(\boldsymbol{p})) / (\text{max}(\boldsymbol{p}) - \text{min}(\boldsymbol{p}))$. Applying the exponential function to \(n(\boldsymbol{p})\) yields non-zero weights $\mathbf{W} = \exp (n(\boldsymbol{v}))$. Similar to~\citet{jiang2021focal} and~\citet{gou2023rethinking}, we introduce hyperparameters $\alpha$ and $\beta$ to enhance the expressibility and controllability of the weights 
\begin{equation}
\mathbf{W} (\boldsymbol{p}; \alpha, \beta) = \alpha \cdot \exp (\beta \cdot n(\boldsymbol{p})). 
\end{equation}

During training, the weighting matrix $\mathbf{W} (\boldsymbol{p}; \alpha, \beta)$ is utilized to adjust the weights. This adjustment can be implemented either as a penalty term in the loss function or through a two-step training approach. Due to limited space, detailed implementations with this learnable frequency-aware loss prior are provided in Appendix~\ref{app:loss_imp}.

\begin{figure}[h!]
\begin{center}
\centerline{\includegraphics[width=\columnwidth]{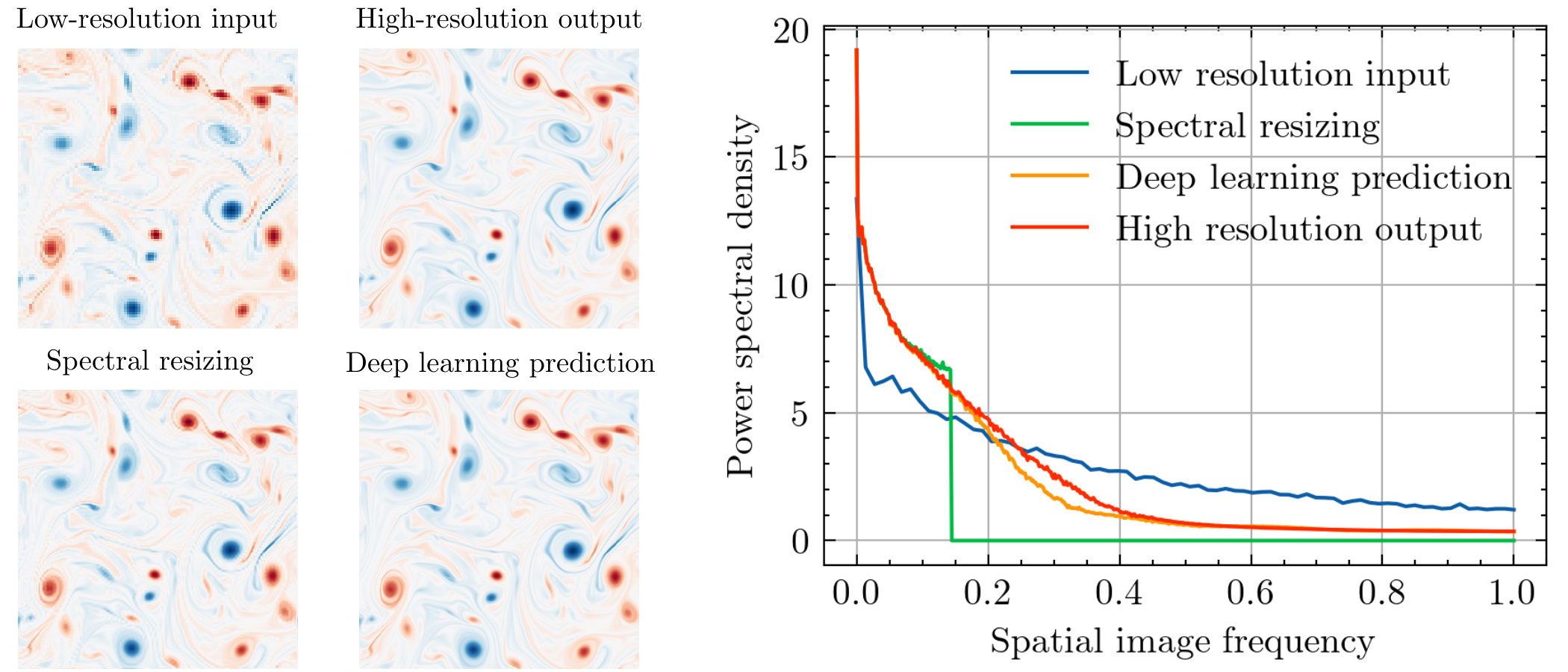}}
\caption{Distinguishing between pixels of different frequency regions in image space poses a challenge. Deep learning predictions often show high visual perception metrics when compared with target HR images (See the left representations). We analyze the images in the frequency domain and observe that the power spectra of HR images and those produced by deep learning models begin to diverge at a certain frequency (e.g., 0.2 in this example). To identify this frequency, spectral resizing is applied to LR inputs, revealing a clear demarcation in frequency regions. This demarcation aligns with the frequency divergence. Building on this, we introduce a static structure prior, created by subtracting low-frequency signals (obtained via spectral resizing) from deep learning predictions in the image space. This prior is then utilized to refine and enhance the network training process.}
\label{fig:loss}
\end{center}
\end{figure}

\begin{table*}[t!]
\centering
\caption{Quantitative comparison results for the $\times 4$ super-resolution task. Superior performance is denoted by a smaller MSE and higher PSNR/SSIM values. Light red indicates the top performer, while light green represents the second-best performer among all baselines and our method for each metric.}
\resizebox{2\columnwidth}{!}{
\begin{tabular}{l c c c c c c c c c c c c}
\hline 
\multicolumn{1}{c}{} & \multicolumn{3}{c}{ Turbulence } & \multicolumn{3}{c}{ Temperature } & \multicolumn{3}{c}{ Kinetic Energy } & \multicolumn{3}{c}{ Water Vapor} \\
\cmidrule(rl){2-4} \cmidrule(rl){5-7} \cmidrule(rl){8-10} \cmidrule(rl){11-13}
Model & MSE & PSNR & SSIM & MSE & PSNR & SSIM & MSE & PSNR & SSIM & MSE & PSNR & SSIM \\
\hline
\multicolumn{13}{c}{ \textit{single-scale} baselines } \\
SRCNN & 1.196e-4 & 39.138 & 0.972 & 5.206e-5 & 42.690 & 0.980 & 3.595e-4 & 33.997 & 0.924 & 9.992e-5 & 39.912 & 0.975 \\
ESPCN & 1.587e-5 & 47.979 & 0.993 & 2.732e-5 & 45.472 & 0.985 & 1.896e-4 & 36.172 & 0.956 & 4.257e-5 & 43.417 & 0.985 \\
EDSR & 7.303e-6 & 51.305 & 0.997 & 2.831e-5 & 45.317 & 0.985 & 1.963e-4 & 36.465 & 0.957 & 4.890e-5 & 42.815 & 0.983 \\
WDSR & \cellcolor{green!15}6.270e-6  & \cellcolor{green!15}51.898 & \cellcolor{green!15}0.997 & 2.441e-5 & 45.984 & 0.986 & 1.769e-4 & 36.792 & 0.955 & 4.837e-5 & 42.662 & 0.983 \\
SwinIR & 2.789e-5 & 45.531 & 0.997 & 2.633e-5 & 45.722 & 0.986 & \cellcolor{green!15}1.736e-4 & \cellcolor{green!15}36.874 & \cellcolor{green!15}0.957 & \cellcolor{green!15}3.209e-5 & \cellcolor{green!15}44.449 & \cellcolor{green!15}0.986 \\
\hline
\multicolumn{13}{c}{ \textit{arbitrary-scale} baselines } \\
MetaSR & 7.988e-5 & 40.879 & 0.976 & 5.270e-5 & 42.639 & 0.978 & 2.199e-4 & 35.751 & 0.946 & 8.489e-5 & 40.564 & 0.977 \\
LIIF & 2.818e-5 & 45.482 & 0.988 & 2.969e-5 & 45.273 & 0.984 & 1.888e-4 & 35.944 & 0.943 & 5.460e-5 & 42.336 & 0.980 \\
LTE & 1.032e-5 & 48.792 & 0.993 & 2.796e-5 & 45.486 & 0.985 & 1.663e-4 & 36.313 & 0.951 & 3.266e-5 & 44.052 & 0.986 \\
DFNO & 2.147e-4 & 36.543 & 0.985 & 1.905e-4 & 37.048 & 0.980 & 2.914e-4 & 34.397 & 0.936 & 3.967e-4 & 33.867 & 0.944 \\
SRNO & 9.083e-6 & 50.403 & 0.996 & \cellcolor{green!15}2.287e-5 & \cellcolor{green!15}46.355 & \cellcolor{green!15}0.987 & 2.083e-4 & 36.635 & 0.949 & 3.446e-5 & 44.299 & 0.986 \\
\hline
HiNOTE & \cellcolor{red!10}6.112e-6 & \cellcolor{red!10}51.903 & \cellcolor{red!10}0.997 & \cellcolor{red!10}1.983e-5 & \cellcolor{red!10}46.424 & \cellcolor{red!10}0.987 & \cellcolor{red!10}1.138e-4 & \cellcolor{red!10}36.995 & \cellcolor{red!10}0.958 & \cellcolor{red!10}3.112e-5 & \cellcolor{red!10}44.501 & \cellcolor{red!10}0.987 \\
\hline
\end{tabular}
}
\label{tab:single_scale_quan}
\end{table*}

\begin{table*}[t!]
\centering
\caption{Quantitative comparison results for the \textit{arbitrary-scale} SR tasks. For each metric, light red indicates the top performer and light green represents the second-best performer.}
\resizebox{2\columnwidth}{!}{
\begin{tabular}{l c c c c c c c c c c c c}
\hline 
\multicolumn{1}{c}{} & \multicolumn{3}{c}{ $\times 4.6$ } & \multicolumn{3}{c}{ $\times 8.2$ } & \multicolumn{3}{c}{ $\times 15.7$ } & \multicolumn{3}{c}{ $\times 32$} \\
\cmidrule(rl){2-4} \cmidrule(rl){5-7} \cmidrule(rl){8-10} \cmidrule(rl){11-13}
Model & MSE & PSNR & SSIM & MSE & PSNR & SSIM & MSE & PSNR & SSIM & MSE & PSNR & SSIM \\
\hline
\multicolumn{13}{c}{ Interpolation Methods } \\
Bilinear & 8.625e-5 & 39.177 & 0.973 & 3.141e-4 & 33.5634 & 0.936 & 8.053e-4 & 29.475 & \cellcolor{green!10}0.913 & \cellcolor{red!10}1.703e-3 & 26.223 & \cellcolor{red!10}0.903  \\
Bicubic & 9.323e-5 & 38.839 & 0.971 & 3.518e-4 & 33.0714 & 0.931 & 9.126e-4 & 28.932 & 0.909 & \cellcolor{green!10}1.960e-3 & 25.612 & \cellcolor{green!10}0.902  \\
Nearest & 1.341e-4 & 37.258 & 0.934 & 4.716e-4 & 31.7989 & 0.859 & 1.179e-3 & 27.817 & 0.831 & 2.454e-3 & 24.635 & 0.841  \\
\hline
\multicolumn{13}{c}{ Deep Learning Methods } \\
MetaSR & 9.274e-5 & 40.295 & 0.974 & 4.043e-4 & 33.900 & 0.904 & 1.227e-3 & 29.077 & 0.842 & 2.716e-3 & 25.628 & 0.806  \\
LIIF & 2.818e-5 & 45.482 & 0.988 & 2.671e-4 & 35.715 & 0.940 & 9.731e-4 & 30.101 & 0.886 & 2.283e-3 & 26.397 & 0.874  \\
LTE & \cellcolor{green!10}7.860e-6 & \cellcolor{green!10}50.985 & \cellcolor{green!10}0.996 & 2.080e-4 & 36.759 & 0.951 & 1.033e-3 & 29.796 & 0.875 & 2.466e-3 & 26.018 & 0.844 \\
DFNO & 2.364e-4 & 36.102 & 0.984 & 4.339e-4 & 33.489 & 0.934 & 1.188e-3 & 29.112 & 0.878 & 2.412e-3 & 26.038 & 0.854  \\
SRNO & 9.272e-6 & 50.081 & 0.995 & \cellcolor{green!10}1.634e-4 & \cellcolor{green!10}37.852 & \cellcolor{green!10}0.959 & \cellcolor{green!10}8.008e-4 & \cellcolor{green!10}30.759 & 0.912 & 2.127e-3 & \cellcolor{green!10}26.708 & 0.879  \\
\hline
HiNOTE & \cellcolor{red!10}7.121e-6 & \cellcolor{red!10}51.225 & \cellcolor{red!10}0.996 & \cellcolor{red!10}8.906e-5 & \cellcolor{red!10}40.002 & \cellcolor{red!10}0.976 & \cellcolor{red!10}7.361e-4 & \cellcolor{red!10}32.164 & \cellcolor{red!10}0.920 & 2.041e-3 & \cellcolor{red!10}26.772 & 0.891 \\
\hline
\end{tabular}
}
\label{tab:arbitrary_scale_quan}
\end{table*}

\begin{figure*}[ht]
\begin{center}
\centerline{\includegraphics[width=2\columnwidth]{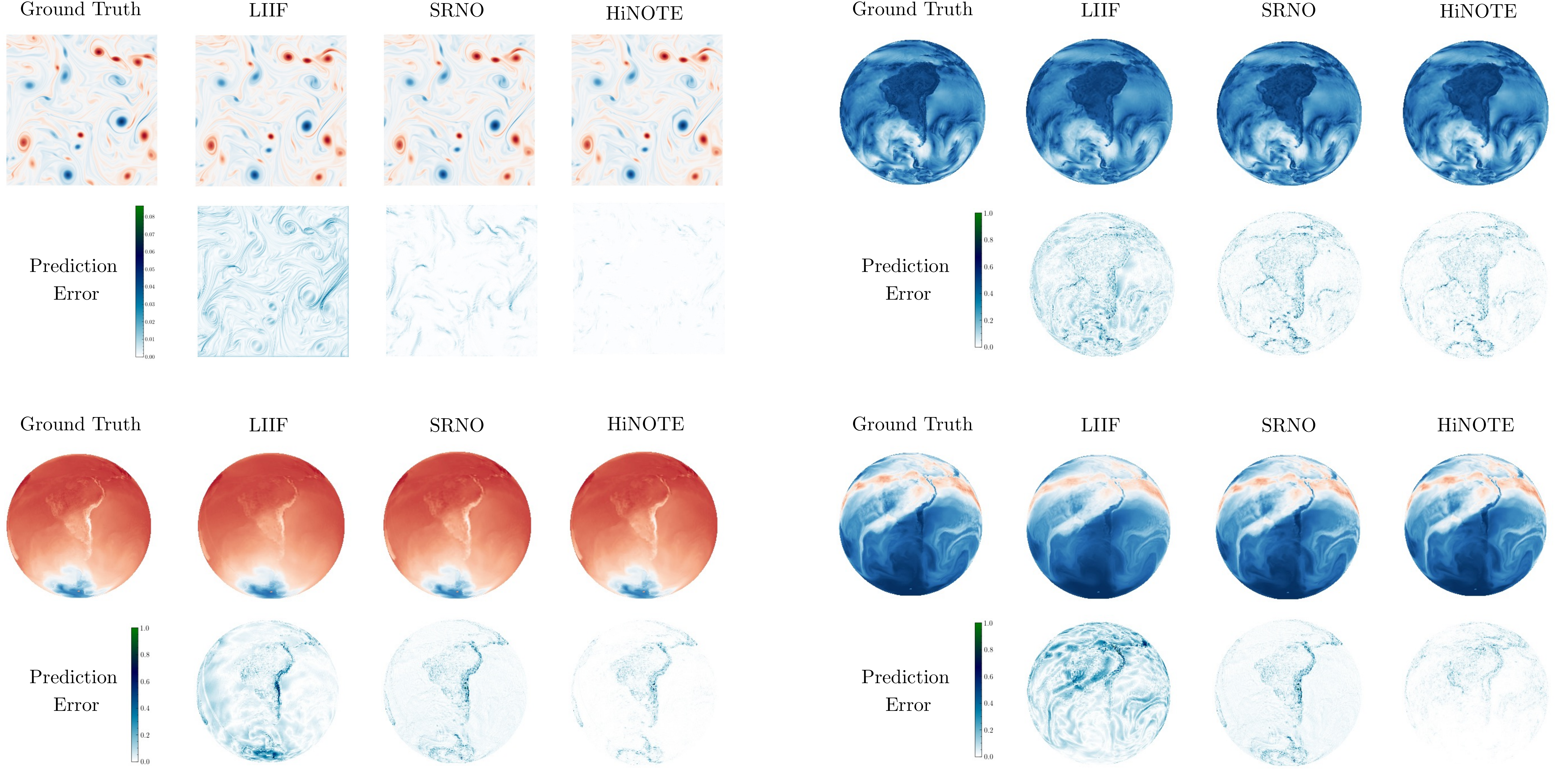}}
\caption{Qualitative comparison with state-of-the-art (SOTA) methods for arbitrary-scale SR. Top-left: turbulence flow; Top-right: kinetic energy; Bottom-left: temperature; and Bottom-right: water vapor.}
\label{fig:qualitative_comparison}
\end{center}
\vskip -0.2in
\end{figure*}


\vspace{-5mm}

\section{Experiments}
\label{experiments}

\subsection{Experimental Setup}
\label{exp_setup}

\textbf{Datasets.} \textcolor{black}{We evaluate our proposed method against baseline methods using four diverse datasets. Detailed information about the generation and preprocessing of these four datasets can be found in Appendix~\ref{app:data}.}

\begin{itemize}
    \item \textbf{Turbulence Flow}: We explore two-dimensional Kraichnan turbulence in a doubly periodic square domain spanning $[0, 2 \pi]^2$. The Navier-Stokes equation is solved through direct numerical simulation to generate the required data~\citep{pawar2023frame}.
    \item \textbf{Global Weather Pattern}: We utilize weather data encompassing ERA5 reanalysis data, high-resolution simulated surface temperature at 2 meters, kinetic energy at 10 meters above the surface, and total column water vapor~\citep{hersbach2020era5}.
    \textcolor{black}{\item \textbf{SEVIR}: We utilize the Storm EVent ImagRy (SEVIR) dataset, which comprises a large, curated collection of labeled examples. This dataset encompasses various weather phenomena, including thunderstorms, convective systems, and related events~\citep{veillette2020sevir}.}
    \textcolor{black}{\item \textbf{MRI}: The magnetic resonance imaging (MRI) dataset includes a variety of snapshots from multiple sources. Our primary focus is on brain scans that utilize a horizontal sampling mask~\citep{jalal2021robust}.}
\end{itemize}

\textbf{Baselines.} We benchmark HiNOTE against several well-acknowledged and advanced models, encompassing five \textit{single-scale} SR baselines: SRCNN~\citep{dong2015image}, ESPCN~\citep{shi2016real}, EDSR~\citep{lim2017enhanced}, WDSR~\citep{yu2018wide}, SwinIR~\citep{liang2021swinir}, and five \textit{arbitrary-scale} SR baselines: MetaSR~\citep{hu2019meta}, LIIF~\citep{chen2021learning}, LTE~\citep{lee2022local}, DFNO~\citep{yang2023fourier}, and SRNO~\citep{wei2023super}. Due to space constraints, the configuration details for each baseline model are provided in Appendix~\ref{app:model}.

\textbf{Evaluation Protocol.} We utilize an A100 GPU with 48GB capacity. Training models directly on HR data poses significant challenges due to memory constraints. Consequently, we employ a strategy of randomly cropping each high-resolution snapshot into smaller segments for training purposes. To ensure fairness in comparison, all methods are trained using the L1 loss function for 300 epochs. This training employs the AdamW optimizer~\citep{loshchilov2018decoupled}, with the initial learning rate determined through hyperparameter tuning. Due to the varying sizes of each model and the limitations imposed by GPU memory, batch sizes may differ (Details are available in Appendix~\ref{app:baseline}). For the evaluation of model performance, we utilize mean squared error (MSE), peak signal-to-noise ratio (PSNR), and structural similarity index (SSIM) as our primary metrics.

\subsection{Main Results}
\label{main_res}

\textbf{Single-scale SR Performance.} In a single-scale SR task ($32 \times 32$ to $128 \times 128$), various models were trained and their performance compared, as shown in Table~\ref{tab:single_scale_quan}. Key observations include: (1) HiNOTE consistently outperforms other models in all SR tasks. It shows notable improvements in weather data, e.g., a $13.29\%$ enhancement on Temperature, a $34.44\%$ improvement on Kinetic Energy and a $3.09\%$ improvement on Water Vapor. 'Improvement' here refers to the relative error reduction compared to the second-best model. (2) Other baselines also show reasonable performance. SwinIR, using a hierarchical transformer architecture and shifted window strategy, is the most competitive across datasets but has a higher computational demand, with 2.1 million parameters compared to HiNOTE's 1.5 million. HiNOTE, with its Galerkin-type self-attention, offers reduced computational needs. (3) Convolutional neural network (CNN)-based models like EDSR and WDSR outperform transformer-based models like SwinIR in turbulence data. On the other hand, HiNOTE leverages a UNet-based hierarchical model architecture, enabling it to effectively learn datasets characterized by patterns spanning various scales. These finding aligns with the nature of turbulence, a small-scale phenomenon, and the strengths of CNNs in capturing fine details, contrasting with attention blocks' effectiveness in larger-scale weather patterns~\citep{gao2022earthformer}. Complete results for the additional two datasets can be found in the Appendix~\ref{app:qual} and~\ref{app:quan}.

\textbf{Arbitrary-scale SR Performance.} Table.\ref{tab:arbitrary_scale_quan} presents a quantitative comparison of HiNOTE with SOTA arbitrary-scale SR models and traditional interpolation methods. For training, all deep learning models employ an upsampling ratio $s_i$ randomly drawn from a uniform distribution $\mathcal{U}[1, 4]$. We tested $s_i$ ranging from a moderate extrapolation of 4.6 to an extreme ratio of 32, noting that these ratios were not included in training. The results reveal two key findings: (1) HiNOTE surpasses current SOTA methods, achieving an average improvement of $20.11\%$ over the second-best model, SRNO, across various upsampling ratios; (2) As $s_i$ increases, the relative performance of deep learning models compared to classical interpolation methods diminishes. Notably, at $s_i = 32$, bilinear interpolation often outperforms deep learning in many metrics. This suggests that the reliable extrapolation range for deep learning models is approximately 16, considering their training on a maximum upsampling ratio of 4. Including larger upsampling ratios in training could potentially extend this range.

\textbf{Showcases.} Fig.\ref{fig:qualitative_comparison} presents qualitative comparisons between the HiNOTE and other leading arbitrary-scale SR methods, specifically LIIF and SRNO. An expanded qualitative comparison is detailed in the Appendix~\ref{app:qual}. The results illustrate HiNOTE's ability to generate super-resolved images with notably sharper textures compared to these methods. For example, in the second row, which displays the absolute error between the target and the predictions, HiNOTE demonstrates superior performance to LIIF, as evidenced by the significantly lower error margins. Against SRNO, the previously best-performing arbitrary-scale SR method, HiNOTE shows marked improvements, especially noticeable in the errors pertaining to ocean areas in the images. This enhanced performance further substantiates the effectiveness of HiNOTE's hierarchical structure in accurately modeling multi-scale data, such as climate-related imagery.

\begin{figure}[h!]
\vskip -0.15in
\begin{center}
\centerline{\includegraphics[width=\columnwidth]{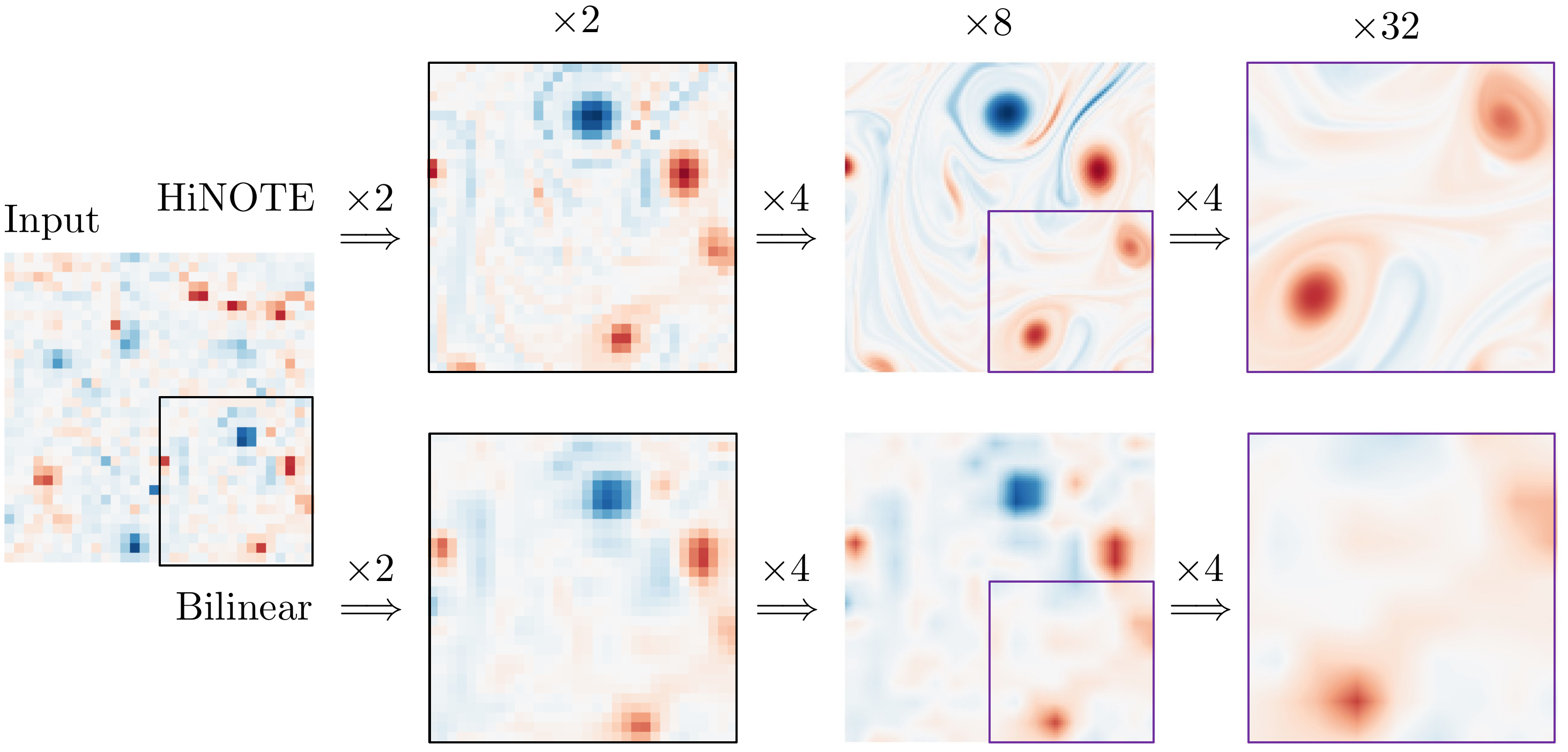}}
\caption{Qualitative demonstration of continuous representation learning: model performance evaluated on two instances randomly chosen from the test dataset, across various upsampling ratios.}
\label{fig:continuous}
\end{center}
\vskip -0.3in
\end{figure}

\textbf{Continuous Representation.} Fig.~\ref{fig:continuous} presents a comparative analysis of arbitrary SR methods: the bilinear interpolation-based approach versus the deep learning-based HiNOTE. Each subplot displays the predicted high-resolution output at different upsampling ratios using turbulence flow data, with specific focus on regions abundant in high-frequency structures (highlighted by rectangular boxes). The comparison reveals the limitations of bilinear interpolation, particularly its restricted context awareness and reliance on neighboring pixels, resulting in imprecise high-frequency detail representation. This deficiency is more pronounced at higher upsampling ratios (e.g., $\times 32$), producing images that are blurry and lack detail. Conversely, HiNOTE effectively reconstructs fine structures, underscoring the significance of global integral kernels in capturing overall structures.

\subsection{Ablation Study}
\label{ablations}


\textbf{Necessity of Refining Features.} To investigate the importance of refining deep features, we trained HiNOTE across various upsampling ratios. Next, performance was evaluated using a test dataset with an upsampling ratio fixed at 8. The results reveal that enhancing the spatial resolution of feature maps improves performance in tasks such as SR. Notably, omitting up-sampling resulted in the lowest performance ($\times 1$). The optimal up-sampling ratio varies with the problem. In this case, a model trained with a fourfold increase outperforms a twofold increase by $4.16\%$.

\begin{table}[h]
\centering
\caption{Experimental results for different up-sampling ratios.}
\begin{tabular}{c c c c}
\hline
Up-sampling ratio & $\times 1$ & $\times 2$ & $\times 4$ \\
\hline
MSE & 9.869e-5 & \cellcolor{green!10}9.307e-5 & \cellcolor{red!10}8.919e-5 \\
\hline
\end{tabular}
\label{tab:ablation_featup}
\end{table}

\textcolor{black}{
\textbf{Efficiency of the Upsampling Module.} It is critical to note that adding an upsampling layer at the end of the network increases the computational cost. Traditionally, a SR model requires a deep encoder to extract complex, abstract features through upsampling at the end of the network. In contrast, HiNOTE incorporates the iterative kernel integral as a central feature, allowing for fewer channels thanks to our proposed patch ensemble approach. Consequently, our encoder requires fewer layers, significantly reducing the overall computational demands. We have performed a comparative analysis to assess the computational effects of including versus excluding the upsampling module. The results, detailed below, demonstrate a minimal increase in computational cost due to the efficient design of our model.
}

\begin{figure}[h!]
\begin{center}
\centerline{\includegraphics[width=0.51\columnwidth]{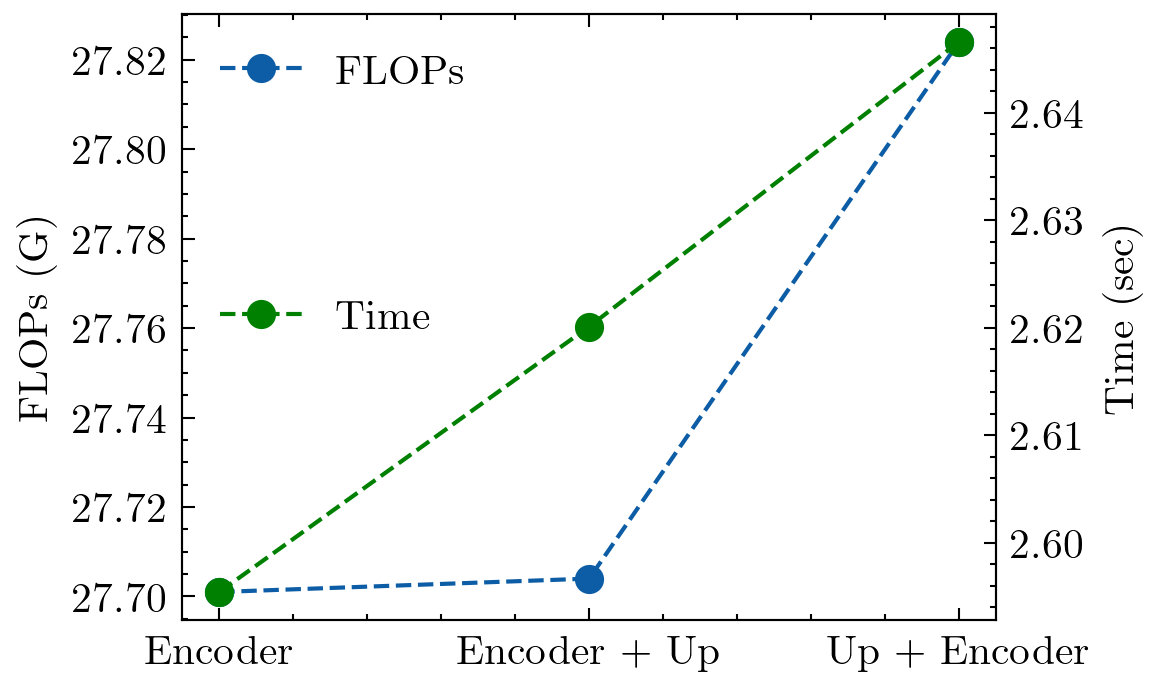}\includegraphics[width=0.49\columnwidth]{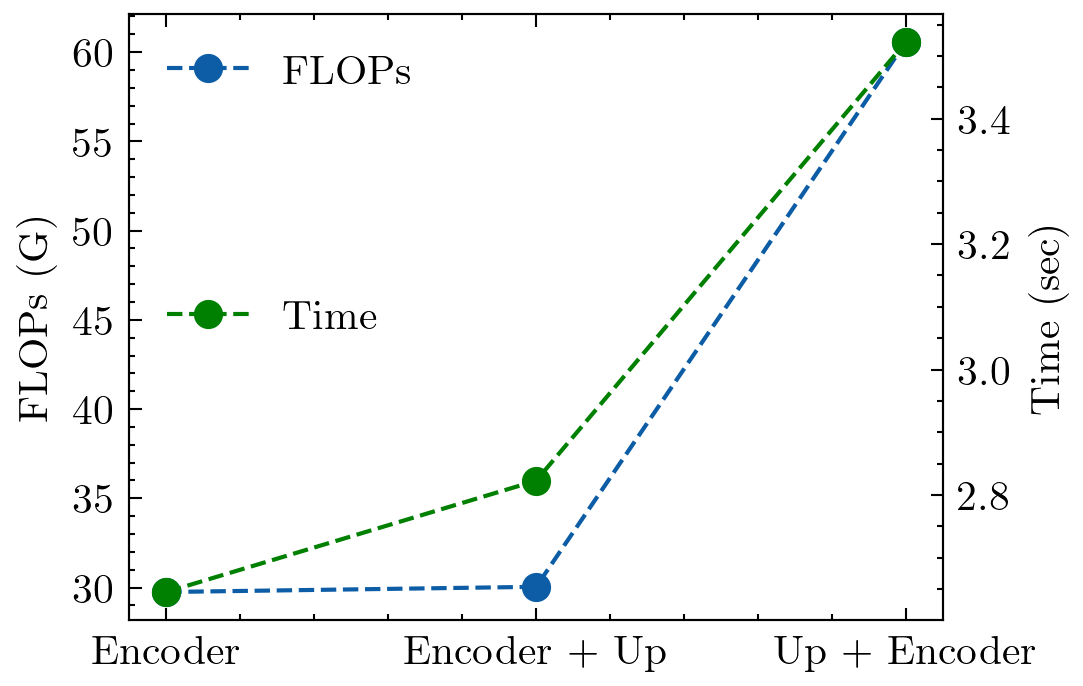}}
\caption{Efficiency comparison of three models: (1) without upsampling, (2) with upsampling after encoder, and (3) with upsampling before encoder.}
\label{fig:upsampling_efficiency}
\end{center}
\vskip -0.3in
\end{figure}

\textcolor{black}{
\textbf{Ablations on Self-attention Mechanisms.} The primary motivation of employing Galerkin-type self-attention is to decrease computational complexity. Among various self-attention mechanisms that also reduce complexity, Galerkin-type self-attention is preferred due to its origin in operator-based learning problems, aligning closely with our model design. Here, we have conducted an ablation study of different self-attention mechanisms: Vanilla~\citep{vaswani2017attention}, FAVOR~\citep{choromanski2020rethinking}, and ProbSparse~\citep{zhou2021informer}. Due to vanilla attention's high computational demands, especially in GPU memory usage, we have scaled down the resolution to $128 \times 128$ for practical evaluation, instead of using the original $1024 \times 1024$ resolution of the turbulence data. FLOPs are calculated per sample. We have observed that while the FAVOR method from the Performer and the ProbSparse self-attention from the Informer significantly reduce computational complexity, adapting these sequence-to-sequence models for computer vision tasks including super-resolution necessitates further optimization, such as adopting patch-based learning similar to ViT~\citep{dosovitskiy2021an}. This adaptation process may influence the computational cost metrics, potentially biasing them. For this reason, the computational cost metrics measured by the wall clock time may be biased due to our naive re-implementation of Performer and Informer not being optimized for this specific context. Despite these considerations, our findings indicate superior performance from the Galerkin-type self-attention. This is likely due to its specific design for operator learning problems. However, recent studies have reported variants of Galerkin-type attention that achieve even better results, which merits further investigation~\citep{hao2023gnot}.
}

\begin{table}[h!]
\centering
\caption{Ablation study for self-attention mechanisms.}
\resizebox{\columnwidth}{!}{
\begin{tabular}{c c c c c}
\hline
Model & Params. (M) & FLOPs (G) & Time (sec) & MSE \\
\hline
Vanilla & \cellcolor{red!10}1.441 & 28.062 & 5.2147 & 3.382e-5 \\
FAVOR & \cellcolor{green!10}1.665 & \cellcolor{green!10}5.622 & 2.4746 & 1.002e-5 \\ 
ProbSparse & 1.681 & 5.644 & \cellcolor{green!10}2.4688 & \cellcolor{green!10}9.941e-6 \\
Galerkin & 1.709 & \cellcolor{red!10}2.708 & \cellcolor{red!10}0.3278 & \cellcolor{red!10}5.201e-6 \\
\hline
\end{tabular}
\label{tab:ablation_attention}
}
\end{table}

\textbf{Enhancement via Loss Prior.} \textcolor{black}{Table.~\ref{tab:ablation_prior} quantitatively summarizes the model performance, as measured by MSE, both with and without the inclusion of the proposed loss function. The experiments have been conducted with the upsampling ratio set at $4$. The ablation study demonstrates that the inclusion of the proposed loss function enhances performance by at least $1.32\%$ across various datasets.}

\begin{table}[h!]
\centering
\caption{Ablation study for learnable frequency-aware loss prior.}
\resizebox{\columnwidth}{!}{
\begin{tabular}{c c c c c}
\hline
Model design & Turbulence & Temperature & Kinetic Energy & Water Vapor \\
\hline
loss prior (-) & 6.194e-6 & 2.061e-5 & 1.455e-4 & 3.169e-5 \\
loss prior (+) & 6.112e-6 & 1.983e-5 & 1.138e-4 & 3.112e-5 \\
improvement & $1.32\%$ & $3.78\%$ & $21.78\%$ & $1.79\%$ \\
\hline
\end{tabular}
\label{tab:ablation_prior}
}
\end{table}

\begin{figure}[h!]
\begin{center}
\centerline{\includegraphics[width=\columnwidth]{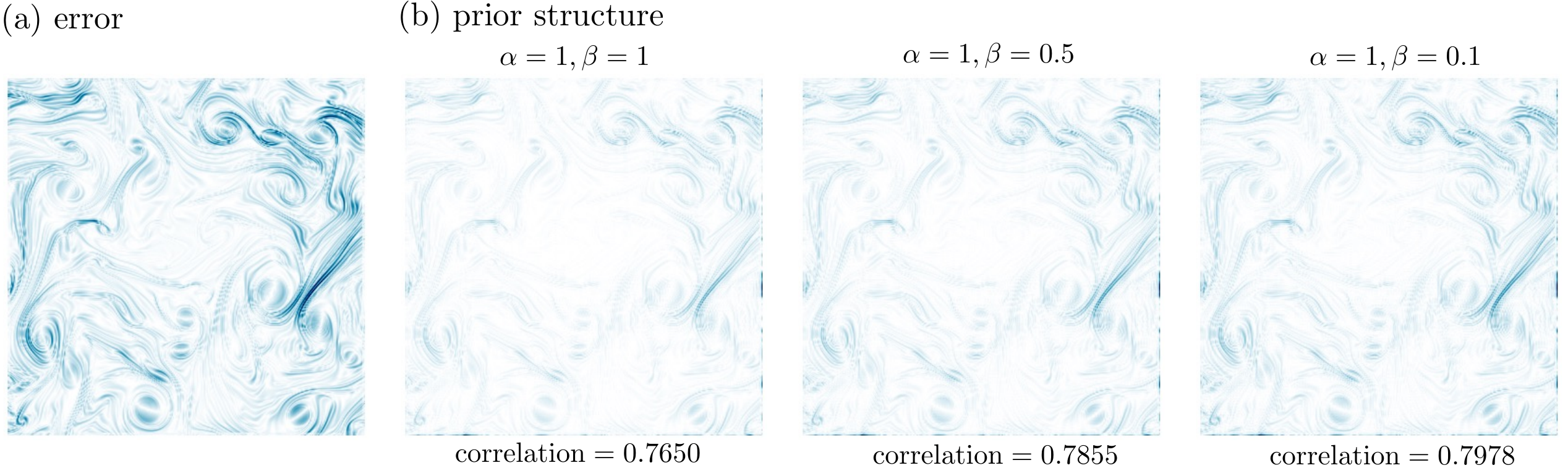}}
\caption{Illustration of the correlation between the high-frequency signals absent in deep learning predictions and the structure priors.}
\label{fig:prior_structure}
\end{center}
\vskip -0.3in
\end{figure}

Furthermore, we investigate the correlation between high-frequency signals missing in deep learning predictions and the structure prior. We begin by calculating the error as the absolute difference between the deep learning predictions and the target. Subsequently, Pearson's correlation coefficient is utilized to assess the relationship between this error and the structure priors, each associated with distinct hyperparameters. In our specific case, we observe that settings of $\alpha=1$ and $\beta=0.1$ yield the most effective structure prior, demonstrating a significant correlation (0.7978) with the error (See Fig.\ref{fig:prior_structure}). This insight allows us to employ this matrix to direct the network towards improved learning of high-frequency signals.

\section{Conclusion}
\label{conclusion}
In this paper, we introduce the \textbf{Hi}erarchical \textbf{N}eural \textbf{O}perator \textbf{T}ransform\textbf{E}r (HiNOTE) for arbitrary-scale super-resolution. HiNOTE conceptualizes digital data as continuous functions, learning mappings between finite-dimensional function spaces, which enables training and generalization across various discretization levels. The process begins by transforming discretized low-resolution input into enhanced spatial-resolution feature maps. Subsequently, these maps are elevated to a higher-dimensional feature space by increasing the channel size to reveal latent features omitted in the original representation. This is followed by applying a hierarchical and self-attention-based kernel approximation before the final output mapping. HiNOTE demonstrates consistent state-of-the-art performance on both turbulence and weather data benchmarks, benefiting from its innovative architecture.


\section*{Acknowledgement} This research were supported by funding from the Advanced Scientific Computing Research program in the Department of Energy’s Office of Science under project $B\&R\#KJ0402010$.

\section*{Broader impact of the work}
\textbf{Broader impacts} The proposed model carries significant broader impacts across various domains, enhancing our ability to interpret and utilize data at unprecedented levels of detail. For instance, in environmental science, it can enhance satellite imagery resolution, enabling more precise climate models and improved decision-making for natural disaster response. In healthcare, it transforms medical imaging by offering detailed visualization of small anatomical features, which could facilitate earlier disease detection. Ultimately, the development of arbitrary-scale super-resolution methods promises to significantly impact society by improving our understanding and management of complex systems, from global ecosystems to human health.

\textbf{Fairness and ethic issues} Our research is committed to upholding ethical standards, focusing on developing models that ensure fairness, reduce biases, and protect privacy. We emphasize transparency in our methods and are willing to share our results to encourage ethical evaluation and peer review.

\nocite{langley00}

\bibliography{example_paper}
\bibliographystyle{icml2024}

\newpage
\appendix
\onecolumn

\section{Data}
\label{app:data}

\subsection{Turbulent Flow}
Turbulent flow is a fluid motion marked by irregular fluctuations in velocity and pressure. In this type of flow, fluid particles move chaotically, causing rapid changes in velocity and direction. The understanding of turbulent flow relies on the Navier-Stokes (NS) equations, which serve as a fundamental framework for studying fluid dynamics~\citep{holmes2012turbulence}. However, solving these equations becomes challenging when turbulence is present, as they couple the velocity field to pressure gradients:

\begin{equation}
\nabla \cdot \mathbf{u}=0, \quad \frac{\partial \mathbf{u}}{\partial t}+\mathbf{u} \cdot \nabla \mathbf{u}=-\frac{1}{\rho} \nabla \mathbf{p}+\nu \nabla^2 \mathbf{u}
\end{equation}

The variables $\mathbf{u}$ and $\mathbf{p}$ represent the velocity field and pressure, respectively. $\rho$ and $\nu$ stand for density and viscosity. In this study, we focus on two-dimensional Kraichnan turbulence in a doubly periodic square domain within $[0, 2 \pi]^2$. The spatial domain is discretized using $2048^2$ degrees of freedom. Solution variables of the NS equations are obtained through direct numerical simulation. A second-order energy-conserving Arakawa scheme computes the nonlinear Jacobian~\citep{arakawa1997computational}, and a second-order finite-difference scheme is employed for the Laplacian of the vorticity~\citep{ren2023superbench}.

\subsection{Global Weather Pattern}
Global weather patterns depict the dominant atmospheric conditions and circulation features defining the Earth's climate globally. These patterns arise from interactions among elements of the Earth's atmosphere, including air temperature, pressure, humidity, and wind. These interactions span spatial and temporal scales over $\mathcal{O}(10)$ orders of magnitude, ranging from micrometers to planetary scales. In this study, we employ ERA5, an advanced atmospheric reanalysis dataset~\citep{hersbach2020era5}. Specifically, ERA5 has global coverage, ranging from the surface to the stratosphere, with a spatial resolution of 0.25 degrees (approximately 25 kilometers). When represented on a cartesian grid, these variables form a $720 \times 1440$ pixel field at any given altitude. Vertically, it is resolved into 37 pressure levels, offering detailed insights up to about 100 kilometers in altitude. The dataset provides hourly estimates of various atmospheric variables, facilitating detailed analyses of short-term weather events. Spanning from 1979 to near-real-time, ERA5 delivers a continuous, long-term record of Earth's atmospheric state. Its creation involves assimilating observations from diverse sources, such as satellite data, ground-based measurements, and meteorological observations, using an advanced numerical model. In this work, we employed three key variables: (1) kinetic energy at 10 meters above the surface, (2) surface temperature at 2 meters, and (3) total column water vapor. The variables are sampled daily at a frequency of 24 hours, spanning a period of 7 years.

\subsection{SEVIR}
\textcolor{black}{The Storm EVent ImagRy (SEVIR)~\citep{veillette2020sevir} dataset is specifically curated to facilitate the development and evaluation of machine learning models in meteorology, with a particular emphasis on nowcasting severe weather events. SEVIR offers a substantial and well-organized collection of labeled examples depicting a range of weather phenomena, including thunderstorms, convective systems, and other related events. Key features of the SEVIR dataset include: (1) High-resolution Imagery: SEVIR encompasses high-resolution spatial and temporal satellite and radar imagery, capturing the intricate dynamics of storm development and progression; (2) Multimodal Data: The dataset incorporates data from various sources, such as satellite imagery (both visible and infrared), radar data, lightning maps, and derived products like Vertically Integrated Liquid (VIL) maps. This multimodal integration enables a comprehensive understanding of storm structures; (3) Event-based Sampling: SEVIR adopts an event-based sampling approach, concentrating on specific storm events. This method provides a focused dataset for the analysis of severe weather, including images captured before, during, and after significant events, which facilitates the temporal analysis of storm evolution; and (4) Wide Coverage: The dataset spans a broad geographic area, primarily across the continental United States, which experiences a diverse array of severe weather events. This extensive coverage enhances the general applicability of models trained on this data.}

\subsection{Data Summary}

Overall, Table.~\ref{tab:dataset} summarizes the datasets utilized in our experiments. The spatial resolution of these datasets has been carefully selected to enable efficient training without requiring multi-GPU computing resources.

\begin{table}[h]
\centering 
\caption{Summary of experiment benchmarks. Turbulence data is simulated using a time step of $\Delta t = 5 \times 10^{-4}$, while weather data is gathered at a sampling frequency of 24 hours.}
\begin{tabular}{c c c c}
\hline
Datasets & Spatial & Temporal & Train/Valid/Test \\
\hline
Turbulence & $1024 \times 1024$ & $0 \rightarrow 4$ & 700/200/100 \\
Weather & $720 \times 1440$ & $2007 \rightarrow 2012$ & 700/200/100 \\
SEVIR & $768 \times 768$ & $2018$ & 700/200/100 \\
MRI & $512 \times 512$ & N/A & 700/200/100 \\
\hline
\end{tabular}
\label{tab:dataset}
\end{table}

\section{Model}
\label{app:model}

\subsection{Baselines}
\label{app:baseline}
In this section, we provide additional training details for all baseline models. To ensure the reliability and consistency of our findings, each experiment was thoughtfully replicated five times. The models were implemented in PyTorch and experiments were conducted on an A100 GPU with 48GB.

\begin{itemize}
    \item \textbf{SRCNN}:~\citet{dong2015image} pioneered the use of a fully convolutional neural network for image super-resolution (SR), enabling end-to-end learning of the LR-to-HR mapping with minimal preprocessing. In our study, we employed SRCNN as a baseline for comparison and followed its default network design. To enhance training, we replaced the original Stochastic Gradient Descent (SGD) optimizer with ADAM~\citep{KingBa15}, balancing convergence speed and stability with a learning rate of $1 \times 10^{-3}$. Regularization was implemented using a weight decay factor of $1 \times 10^{-5}$ to prevent overfitting and encourage generalization. Our training regimen extended over $200$ epochs, with a batch size of $32$ chosen for computational efficiency.
    \item \textbf{ESPCN}:~\citet{shi2016real} introduce an innovative resolution enhancement approach through pixel-shuffle, facilitating deep neural network training within the low-resolution latent space. The study employs the default network architecture, with a learning rate of $1 \times 10^{-3}$, a batch size of $32$, and a weight decay of $1 \times 10^{-4}$. Training spans $200$ epochs, employing the Adam optimizer~\citep{KingBa15}.
    \item \textbf{EDSR}: Utilizing a deep residual network architecture with an extensive array of residual blocks, EDSR~\citep{lim2017enhanced} effectively acquires the LR-to-HR image mapping while capturing hierarchical features. Our comparison study adheres to EDSR's default network configuration, employing 16 residual blocks with a hidden channel size of 64. For optimization, we set the learning rate to $1 \times 10^{-4}$ and incorporate a weight decay of $1 \times 10^{-5}$. The batch size is fixed at $64$, and training spans $300$ epochs, facilitated by the ADAM optimizer~\citep{KingBa15}. 
    \item \textbf{WDSR}:~\citet{yu2018wide} introduced WDSR to enhance reconstruction accuracy and computational efficiency by considering wider features before ReLU in residual blocks. They presented two architectures, WDSR-A and WDSR-B, with WDSR-B being deeper and more powerful but demanding greater computational resources. Our implementation employs 18 lightweight residual blocks with wide activation and a hidden channel of 32. Training utilizes a learning rate of $1 \times 10^{-4}$ and a weight decay of $1 \times 10^{-5}$ over 300 epochs with the ADAM optimizer~\citep{KingBa15}. The batch size is set to 32.
    \item \textbf{SwinIR}: SwinIR~\citep{liang2021swinir} is built upon the sophisticated Swin Transformer architecture~\citep{liu2021swin}, leveraging its capabilities for local attention and cross-window interaction. Key architectural parameters include the use of 6 residual Swin Transformer blocks (RSTB), 6 Swin Transformer layers (STL), a window size of 8, a channel number of 180, and 6 attention heads. For training, a learning rate of $1 \times 10^{-4}$ and a weight decay of $1 \times 10^{-5}$ are selected. The batch size for the training process is configured as $32$.
    \item \textbf{MetaSR}: The Meta-Upscale Module~\citep{hu2019meta} is composed of a stack of $16$ residual blocks. In the encoding phase, we aim to extract $64$ features. During the training phase, we have chosen a learning rate of $1 \times 10^{-4}$ along with a weight decay of $1 \times 10^{-5}$. The training process is conducted with a batch size of 32.
    \item \textbf{LIIF}: Following the original approach~\citep{chen2021learning}, we utilize patches as inputs for the encoder. Denoting the batch size as $B$, we start by randomly selecting $B$ scaling factors ($r_{1 \sim B}$) from a uniform distribution $\mathcal{U}[1, 4]$. Then, we extract $B$ patches from training images, each sized at $\{32 r_i \times 32 r_i\}_i^B$, while their down-sampled counterparts remain $32 \times 32$. For the ground-truth data, we convert these images into pixel samples, with $1024$ samples sampled from each image to ensure consistent shapes within a batch. The encoder, denoted as $E(\cdot)$, is based on EDSR-baseline but excludes its up-sampling modules, generating a feature map of the same size as the input image. The decoding function, denoted as $f(\cdot)$, is implemented as a $5$-layer MLP with ReLU activation functions and hidden dimensions of $256$.
    \item \textbf{LTE}: Lee~\citep{lee2022local} introduced the Local Texture Estimator (LTE), a dominant-frequency estimator designed for natural images. This estimator allows an implicit function to capture fine details during the continuous reconstruction of images. When integrated with a deep super-resolution (SR) architecture, the LTE effectively characterizes image textures in 2D Fourier space. We configure LTE analogously to LIIF. The amplitude and frequency estimators are constructed using $3 \times 3$ convolutional layers, each with 256 output channels. This configuration is equivalent to a fully connected layer when the feature maps are unfolded. The phase estimator consists of a single fully connected layer with a hidden dimension of $128$.
    \item \textbf{DFNO}: Consistent with the original paper~\citep{yang2023fourier}, our implementation follows a specific architectural design: (1) Encoder: Modeled as a residual convolutional network inspired by super-resolution GAN generators~\citep{wang2018esrgan}; (2) Decoder: Implemented as a Fourier Neural Operator; and (3) Upsampling: Achieved using bicubic interpolation. Specifically, the encoder consists of five residual blocks, and the Fourier neural operator comprises four layers of Fourier integral operators with ReLU activation and batch normalization. We built and trained the DFNO model on a turbulence and weather dataset with a $\times 4$ upsampling factor ($32 \times 32$ to $128 \times 128$). Subsequently, we evaluated its performance using various upsampling factors.
    \item \textbf{SRNO}: The SRNO network~\citep{wei2023super} architecture features a key component, namely a feature encoder denoted as $E(\cdot)$. To construct this encoder, we adopt the EDSR-baseline architecture and exclude its upsampling layers, while maintaining output channel dimensions at $d_e = 64$. We also include a multi-head attention mechanism, which involves dividing queries, keys, and values into $n_h$ segments, each with a dimension of $d_z/n_h$. In our specific implementation, we set the embedding dimension as $d_z = 256$ and the number of heads as $n_h = 16$, resulting in 16-dimensional output values from the attention mechanism. The kernel integral operator is applied twice.
\end{itemize}

\subsection{Our Approach}

\subsubsection{Improvements}

\begin{itemize}
    \item \textbf{Feature Refining} In contemporary super-resolution methods, the upsampling layer is typically positioned towards the end of the network. We suggest that enhancing feature maps at the beginning of the network is more effective for SR tasks. 
    \item \textbf{Neural Operator} Implicit neural representations, often parameterized by multi-layer perceptrons (MLPs), struggle with high-frequency information learning due to their point-wise spatial behavior, a phenomenon known as spectral bias. To address this, we substitute the MLP-based inference network with a neural operator. This operator treats images as continuous functions instead of 2D pixel arrays, where each image is a discretization of an underlying function.
\end{itemize}

\begin{figure}[ht]
\begin{center}
\centerline{\includegraphics[width=\columnwidth]{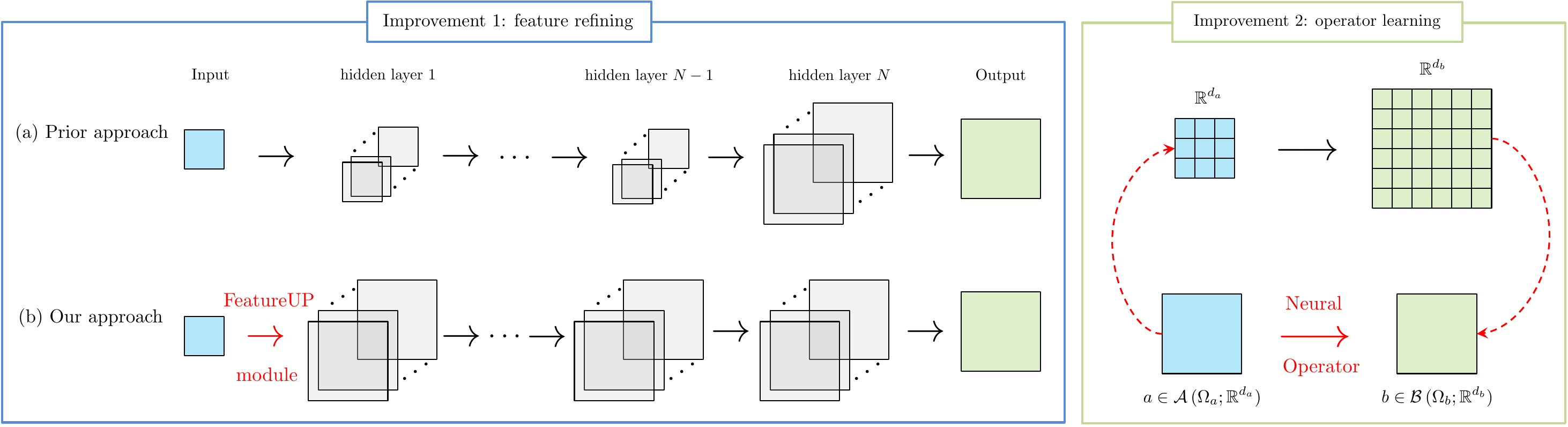}}
\caption{Comparison of the proposed model's enhancements over the previous approach.}
\label{fig:improvment}
\end{center}
\end{figure}

\subsubsection{Loss function}
\label{app:loss_imp}

We propose two approaches to incorporate the loss prior into training our HiNOTE model: (1) utilizing a focal loss-based technique, and (2) employing a two-stage network training process.

\textbf{Approach 1} Similar to the concept of focal loss~\citep{jiang2021focal,gou2023rethinking}, the loss function can be formulated as 

\begin{equation}
\label{eq:app_focal}
\mathcal{L} = \frac{1}{N} \sum_{j=1}^N \mathcal{L}_j \quad \text{where} \quad \mathcal{L}_j = \mathbf{W} (\boldsymbol{p}^{(j)}; \alpha_p, \beta_p) \times \mathbf{W} (\hat{\boldsymbol{b}}^{(j)}; \alpha_{\hat{\boldsymbol{b}}}, \beta_{\hat{\boldsymbol{b}}}) \times \left|\mathcal{G}_\theta\left(\boldsymbol{a}^{(j)}\right) + \mathcal{R}\left(\boldsymbol{a}^{(j)}\right) - \boldsymbol{b}^{(j)} \right|
\end{equation}

In Eq.~\ref{eq:app_focal}, the prior loss $\mathbf{W} (\boldsymbol{p}^{(j)}; \alpha_p, \beta_p)$ quantifies the discrepancy between the target and the output of spectral resizing, while $\mathbf{W} (\hat{\boldsymbol{b}}^{(j)}; \alpha_{\hat{\boldsymbol{b}}}, \beta_{\hat{\boldsymbol{b}}})$ assesses the difference between the target and the deep learning prediction.

\textbf{Approach 2} In the first stage, we employ the standard training procedure. The problem of learning an operator that approximates the mapping between $\mathcal{A}$ and $\mathcal{B}$ naturally takes the form of an empirical risk minimization problem
\begin{equation}
\theta^{\dagger} = \arg \min _{\theta \in \Theta} \mathbb{E}_{\mathbf{a}}\left[\mathcal{C}\left(\mathcal{G}(\mathbf{a}, \theta), \mathcal{G}^{\dagger}(\mathbf{a})\right)\right] \approx \arg \min _{\theta \in \Theta} \frac{1}{N} \sum_{j=1}^N\left\|\boldsymbol{b}^{(j)}-\mathcal{G}_\theta\left(\boldsymbol{a}^{(j)}\right)\right\|_{\mathcal{B}}^2
\end{equation}

where function $\mathcal{C}: \mathcal{B} \times \mathcal{B} \rightarrow \mathbb{R}$ serves as a cost functional, quantifying the distance within the space $\mathcal{B}$. The optimized parameter values from stage 1 serve as the initial values for stage 2, aiming to solve
\begin{equation} 
\arg \min _{\theta \in \Theta} \frac{1}{N} \sum_{j=1}^N \mathbf{W} (\boldsymbol{p}^{(j)}; \alpha, \beta) \cdot \left\|\boldsymbol{b}^{(j)}-\mathcal{G}_\theta\left(\boldsymbol{a}^{(j)}\right)\right\|_{\mathcal{B}}^2 
\end{equation} 

In stage 2, the model prioritizes learning from challenging-to-fit pixels during the initial fitting stage. 


\subsubsection{Hyperparameters}
In deep learning, the design of network architectures and the optimization of hyperparameters are substantial challenges, typically addressed through empirical, problem-specific methods. To facilitate a fair comparison, we have executed extensive hyperparameter tuning and architecture searches for each model under consideration. The specific model configurations for our proposed method are outlined in Table~\ref{tab:model_config}.

\begin{table}[h]
\centering
\caption{Model configurations, hyperparameters, and associated ranges.}
\begin{tabular}{l c}
\hline 
Hyperparameters & Values \\
\hline
Upsample ratio & $\{ \times 1, \times 2, \times 4 \}$ \\
Nonlinear activation & $\{$ ReLU, LeakyReLU, ELU, SELU, GELU, RReLU $\}$ \\
Attention width & $\{ 128, 160, 192, 224, 256 \}$ \\
Attention head & $\{ 2, 4, 8, 16, 32 \}$ \\
\hline
\end{tabular}
\label{tab:model_config}
\end{table}

\subsection{Additional qualitative results}
\label{app:qual}

We present the test predictions of HiNOTE alongside various \textit{arbitrary-scale} super-resolution baseline methods. For each figure, the super-resolved predictions are depicted in the top row, while the bottom row features error maps with respect to the reference data, with darker pixels indicating greater errors. This visual comparison not only highlights the accuracy of our model in generating high-resolution predictions but also provides a quantitative assessment of its performance by visualizing the error distribution across the domain.

\begin{figure}[h!]
\begin{center}
\centerline{\includegraphics[width=0.9\columnwidth]{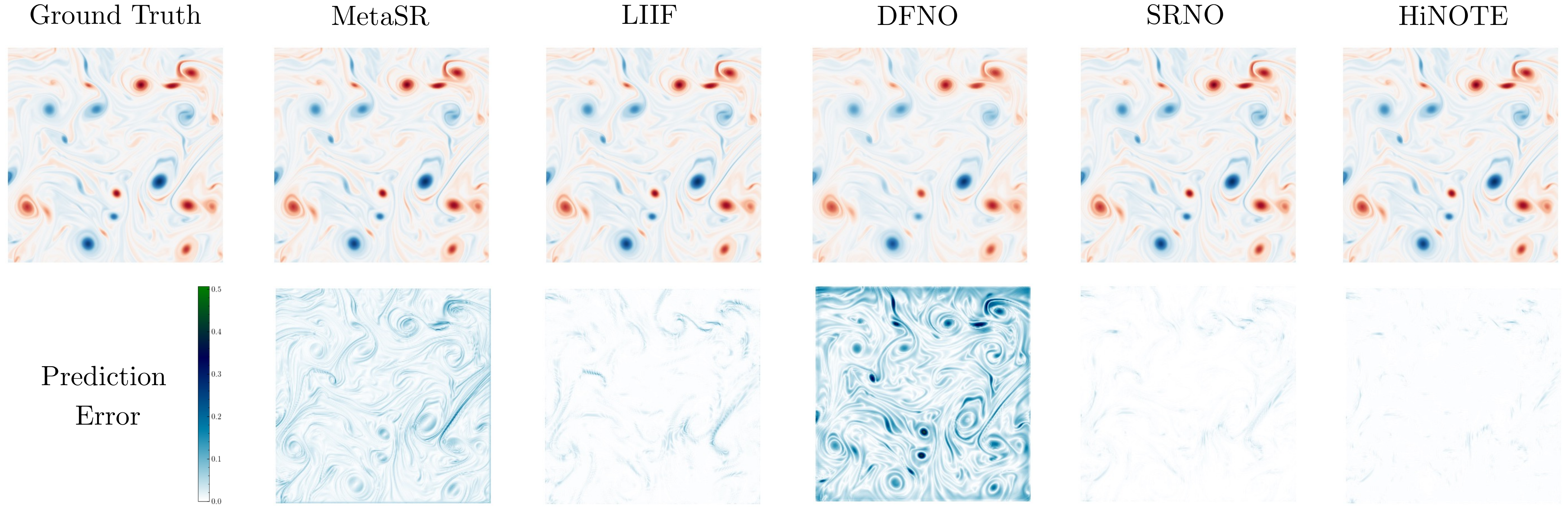}}
\caption{Qualitative assessment of the models on turbulence flow data.}
\label{fig:qual_flow}
\end{center}
\end{figure}

\newpage

\begin{figure}[ht]
\vskip 0.3in
\begin{center}
\centerline{\includegraphics[width=0.9\columnwidth]{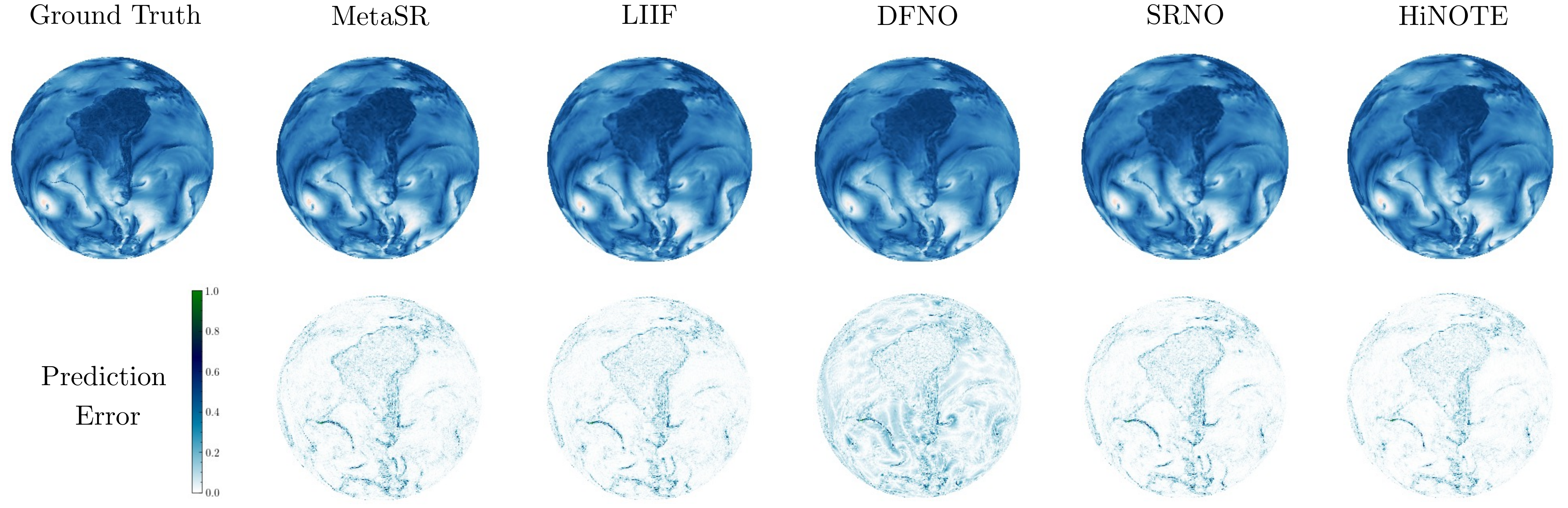}}
\caption{Qualitative assessment of the models on kinetic energy data.}
\label{fig:qual_app2}
\end{center}
\end{figure}

\begin{figure}[h!]
\begin{center}
\centerline{\includegraphics[width=0.9\columnwidth]{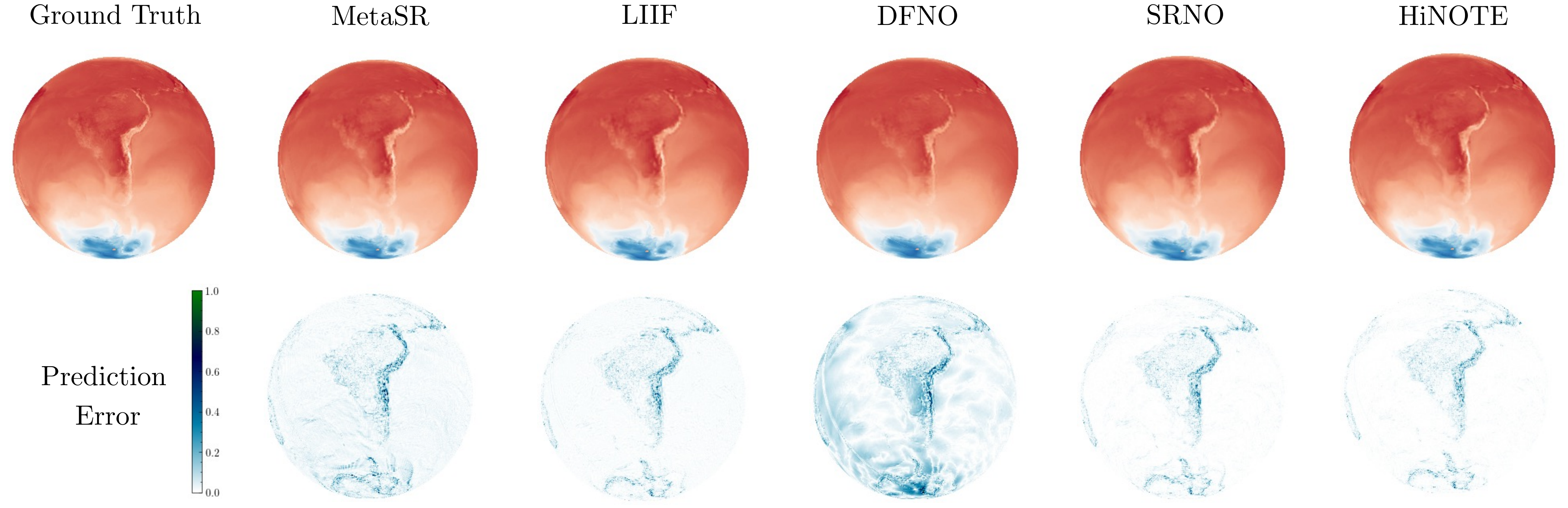}}
\caption{Qualitative assessment of the models on temperature data.}
\label{fig:qual_app3}
\end{center}
\end{figure}

\begin{figure}[h!]
\begin{center}
\centerline{\includegraphics[width=0.9\columnwidth]{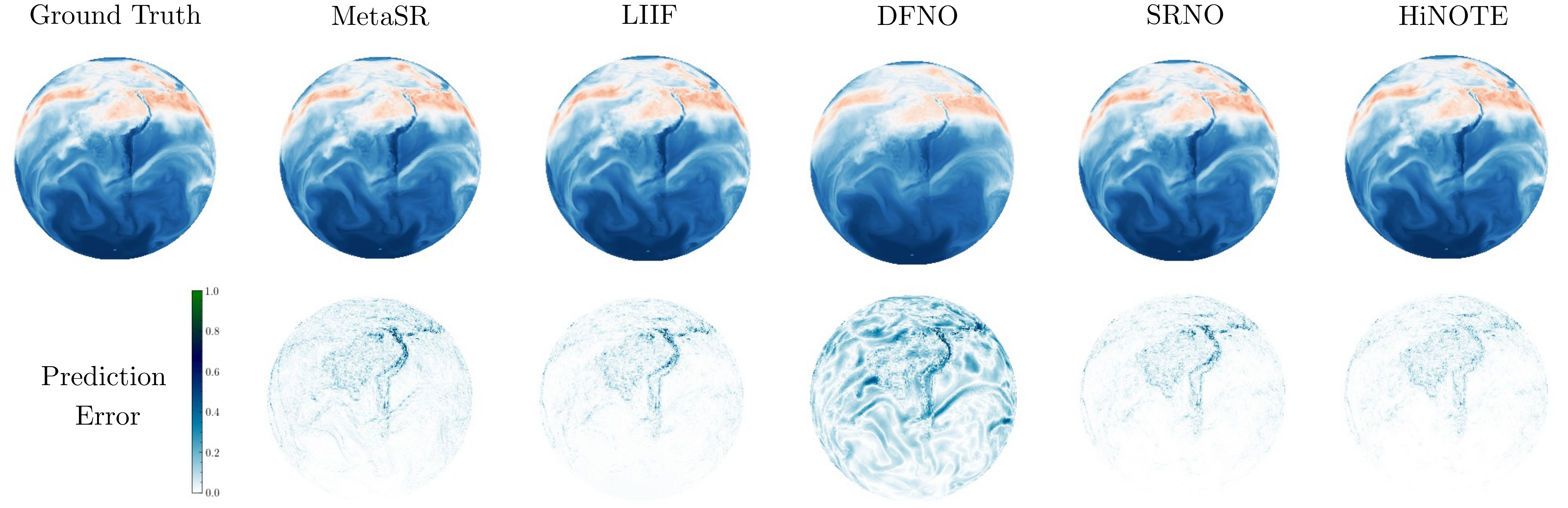}}
\caption{Qualitative assessment of the models on water vapor data.}
\label{fig:qual_app4}
\end{center}
\end{figure}

\begin{figure}[h!]
\begin{center}
\centerline{\includegraphics[width=0.9\columnwidth]{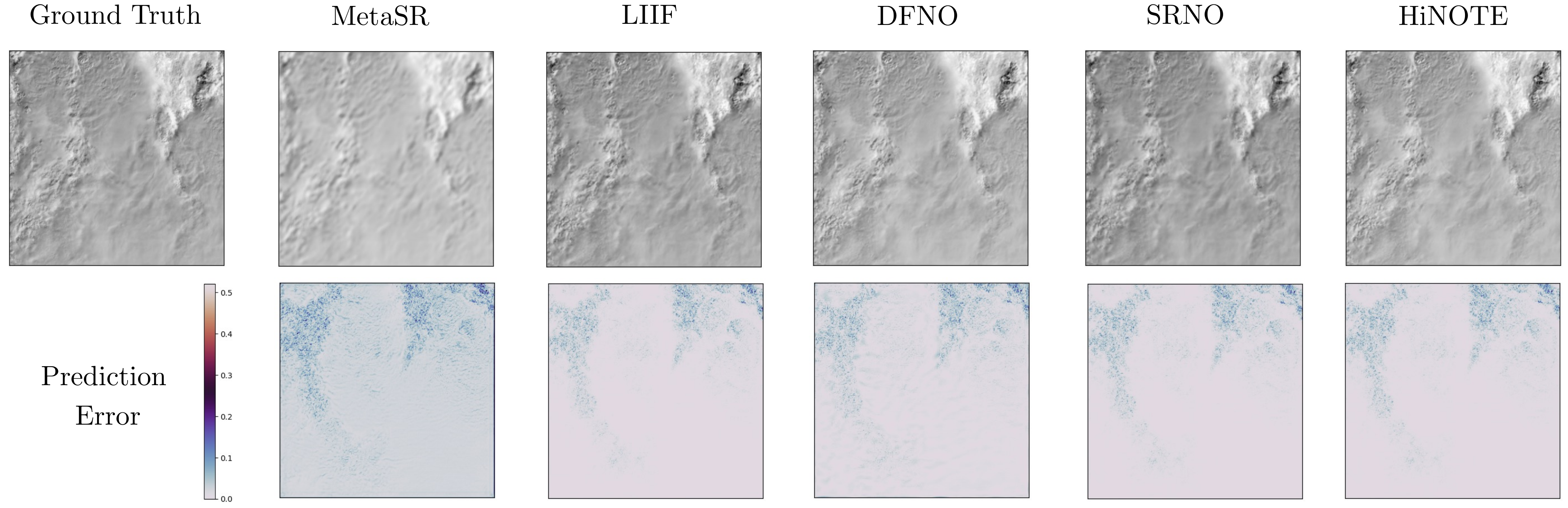}}
\caption{Qualitative assessment of the models on SEVIR data.}
\label{fig:qual_app4}
\end{center}
\end{figure}

\begin{figure}[h!]
\begin{center}
\centerline{\includegraphics[width=0.9\columnwidth]{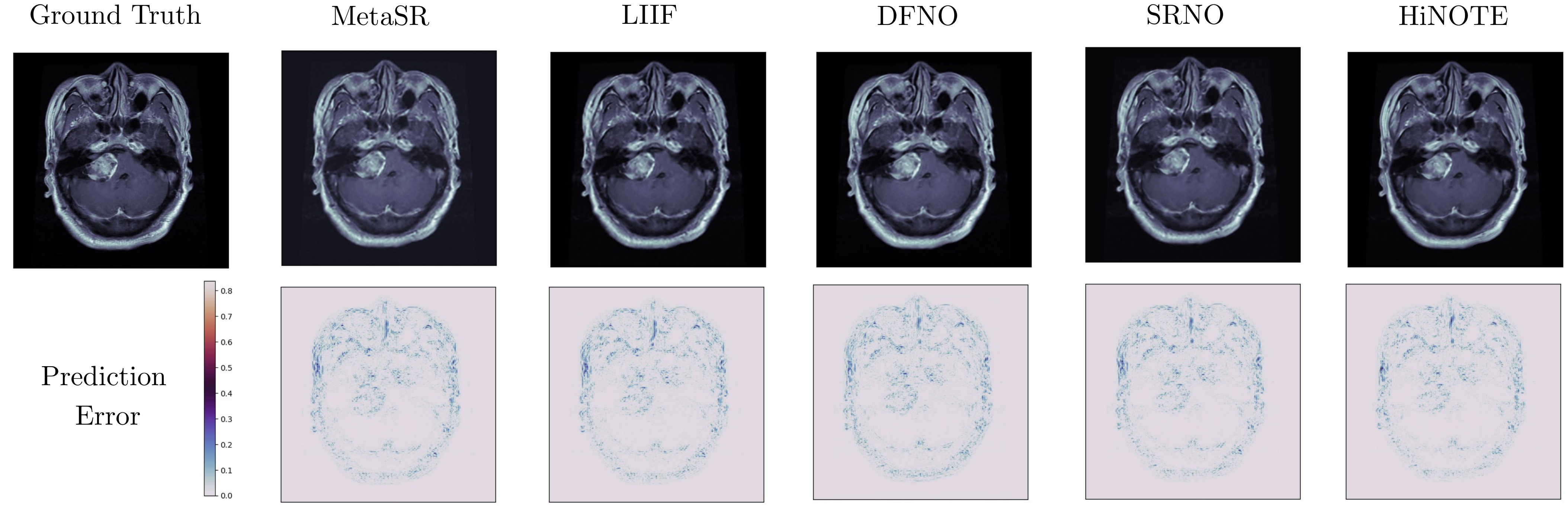}}
\caption{Qualitative assessment of the models on magnetic resonance imaging data.}
\label{fig:qual_app4}
\end{center}
\end{figure}

\subsection{Additional quantitative results}
\label{app:quan}
Here, we present comprehensive quantitative results for the arbitrary-scale super-resolution models applied to the SEVIR and MRI datasets.

\begin{table*}[h]
\centering
\caption{Quantitative comparison results for the \textit{arbitrary-scale} SR tasks on the SEVIR data.}
\resizebox{1\columnwidth}{!}{
\begin{tabular}{l c c c c c c c c c c c c}
\hline 
\multicolumn{1}{c}{} & \multicolumn{3}{c}{ $\times 4.6$ } & \multicolumn{3}{c}{ $\times 8.2$ } & \multicolumn{3}{c}{ $\times 15.7$ } & \multicolumn{3}{c}{ $\times 32$} \\
\cmidrule(rl){2-4} \cmidrule(rl){5-7} \cmidrule(rl){8-10} \cmidrule(rl){11-13}
Model & MSE & PSNR & SSIM & MSE & PSNR & SSIM & MSE & PSNR & SSIM & MSE & PSNR & SSIM \\
\hline
MetaSR & 5.630e-4 & 32.378 & 0.865 & 4.475e-3 & 23.376 & 0.675 & 9.420e-3 & 20.143 & 0.655 & 1.362e-2 & 18.542 & 0.685  \\
LIIF & 1.099e-4 & 39.473 & \cellcolor{green!10}0.965 & 4.059e-4 & 33.799 & 0.923 & 8.763e-4 & 30.457 & 0.894 & 1.504e-3 & 28.110 & 0.879  \\
DFNO & 1.729e-4 & 37.504 & 0.957 & 4.606e-4 & 33.251 & 0.915 & 9.226e-4 & 30.234 & 0.895 & 1.423e-3 & 28.349 & 0.881  \\
SRNO & \cellcolor{green!10}1.093e-4 & \cellcolor{green!10}39.488 & 0.964 & \cellcolor{green!10}3.312e-4 & \cellcolor{green!10}34.674 & \cellcolor{green!10}0.928 & \cellcolor{green!10}6.730e-4 & \cellcolor{green!10}31.595 & \cellcolor{green!10}0.905 & \cellcolor{green!10}1.119e-3 & \cellcolor{green!10}29.385 & \cellcolor{green!10}0.893  \\
\hline
HiNOTE & \cellcolor{red!10}1.048e-4 & \cellcolor{red!10}39.680 & \cellcolor{red!10}0.969 & \cellcolor{red!10}3.191e-4 & \cellcolor{red!10}34.845 & \cellcolor{red!10}0.931 & \cellcolor{red!10}6.302e-4 & \cellcolor{red!10}31.889 & \cellcolor{red!10}0.911 & \cellcolor{red!10}9.981e-4 & \cellcolor{red!10}29.892 & \cellcolor{red!10}0.902 \\
\hline
\end{tabular}
}
\label{tab:SEVIR}
\end{table*}

\vspace{10mm}

\begin{table*}[h]
\centering
\caption{Quantitative comparison results for the \textit{arbitrary-scale} SR tasks on the SEVIR data.}
\resizebox{1\columnwidth}{!}{
\begin{tabular}{l c c c c c c c c c c c c}
\hline 
\multicolumn{1}{c}{} & \multicolumn{3}{c}{ $\times 4.6$ } & \multicolumn{3}{c}{ $\times 8.2$ } & \multicolumn{3}{c}{ $\times 15.7$ } & \multicolumn{3}{c}{ $\times 32$} \\
\cmidrule(rl){2-4} \cmidrule(rl){5-7} \cmidrule(rl){8-10} \cmidrule(rl){11-13}
Model & MSE & PSNR & SSIM & MSE & PSNR & SSIM & MSE & PSNR & SSIM & MSE & PSNR & SSIM \\
\hline
MetaSR & 4.761e-4 & 33.222 & \cellcolor{green!10}0.942 & 3.499e-3 & 24.559 & 0.784 & 9.841e-3 & 20.069 & 0.680 & 1.853e-2 & 17.320 & 0.624  \\
LIIF  & 6.348e-4 & 31.973 & 0.916 & 2.327e-3 & 26.331 & 0.815 & 5.773e-3 & 22.385 & 0.734 & \cellcolor{green!10}1.123e-2 & \cellcolor{green!10}19.495 & 0.676  \\
DFNO & 4.937e-4 & 33.065 & 0.928 & 2.284e-3 & 26.412 & 0.830 & 6.161e-3 & 22.103 & 0.754 & 1.166e-2 & 19.330 & 0.663  \\
SRNO & \cellcolor{green!10}4.464e-4 & \cellcolor{green!10}33.502 & 0.937 & \cellcolor{green!10}2.125e-3 & \cellcolor{green!10}26.726 & \cellcolor{green!10}0.839 & \cellcolor{green!10}5.573e-3 & \cellcolor{green!10}22.538 & \cellcolor{green!10}0.759 & 1.176e-2 & 19.293 & \cellcolor{green!10}0.681  \\
\hline
HiNOTE & \cellcolor{red!10}3.805e-4 & \cellcolor{red!10}34.195 & \cellcolor{red!10}0.947 & \cellcolor{red!10}1.808e-3 & \cellcolor{red!10}27.427 & \cellcolor{red!10}0.871 & \cellcolor{red!10}4.806e-3 & \cellcolor{red!10}23.181 & \cellcolor{red!10}0.786 & \cellcolor{red!10}1.076e-2 & \cellcolor{red!10}19.678 & \cellcolor{red!10}0.704 \\
\hline
\end{tabular}
}
\label{tab:MRI}
\end{table*}


\end{document}